\documentclass[letterpaper, 10 pt, conference]{ieeeconf}  % Comment this line out if you need a4paper

\IEEEoverridecommandlockouts                              % This command is only needed if 
\let\labelindent\relax
\usepackage{times}
\usepackage{epsfig}
\usepackage{graphicx}
\usepackage{enumitem}
\usepackage{amsmath}
\usepackage{amssymb}
\usepackage{hyperref}
\usepackage{amsmath}
\usepackage{amssymb}
\usepackage{multirow}
\usepackage{makecell}
\usepackage{booktabs} 
\usepackage{wrapfig}
\usepackage{gensymb}
\usepackage{xparse} 
\usepackage{caption}
\usepackage{multicol}
\usepackage{subcaption}
\usepackage{caption}
\usepackage[ruled,vlined]{algorithm2e}
\SetKw{Break}{break}

\newcommand{\I}{\mathcal{I}}

\newcommand{\M}{\mathcal{M}}
\newcommand{\F}{\mathcal{F}}
\newcommand{\V}{\mathcal{V}}
\newcommand{\D}{\mathcal{D}}
\newcommand{\cS}{\mathcal{S}}
\newcommand{\R}{\mathbb{R}}

\newcommand{\etal}{\textit{et al}. }
\newcommand{\etc}{\textit{etc}. }

\newcommand{\eg}{\textit{e}.\textit{g}. }

\newcommand{\changed}[1]{}
\setlist{noitemsep,leftmargin=*}

\title{\LARGE \bf
HDMapNet: An Online HD Map Construction and Evaluation Framework
}

\author{Qi Li$^{1,*}$, Yue Wang$^{2,*}$, Yilun Wang$^{3}$ and Hang Zhao$^{1}$ 
\thanks{$^{1}$Qi Li and Hang Zhao are with Tsinghua University, Beijing, China. Hang Zhao is the corresponding author:
{\tt\small hangzhao@mail.tsinghua.edu.cn}}%
\thanks{$^{2}$Yue Wang is with Massachusetts Institute of Technology, MA, USA.}%
\thanks{$^{3}$Yilun Wang is with Li Auto, Beijing, China.}%
}
\begin{document}

\maketitle
\thispagestyle{empty}
\pagestyle{empty}

\renewcommand{\thefootnote}{\fnsymbol{footnote}}
\addtocounter{footnote}{1}
\footnotetext{Equal contribution.}
%%%%%%%%%%%%%%%%%%%%%%%%%%%%%%%%%%%%%%%%%%%%%%%%%%%%%%%%%%%%%%%%%%%%%%%%%%%%%%%%
% \vspace{-10mm}
\begin{abstract}
Constructing HD semantic maps is a central component of autonomous driving.
However, traditional pipelines require a vast amount of human efforts and resources in annotating and maintaining the semantics in the map, which limits its scalability.
In this paper, we introduce the problem of HD semantic map learning, which dynamically constructs the local semantics based on onboard sensor observations. Meanwhile, we introduce a semantic map learning method, dubbed HDMapNet. 
% Meanwhile, we propose a learning-based method, dubbed HDMapNet, to predict and construct the local semantic maps based on sensor observations on-the-go. 
% Our method is immediately applicable to any outdoor birds-eye view semantic estimation tasks.
HDMapNet encodes image features from surrounding cameras and/or point clouds from LiDAR, and predicts vectorized map elements in the bird's-eye view.
We benchmark HDMapNet on nuScenes dataset and show that in all settings, it performs better than baseline methods. Of note, our camera-LiDAR fusion-based HDMapNet outperforms existing methods by more than 50\% in all metrics.
In addition, we develop semantic-level and instance-level metrics to evaluate the map learning performance.
Finally, we showcase our method is capable of predicting a locally consistent map. By introducing the method and metrics, we invite the community to study this novel map learning problem. 
% Code and evaluation kit will be released to facilitate future development.
\end{abstract}

% \keywords{Autonomous Driving, Computer Vision, High-definition Maps} 
% \vspace{-5mm}
\section{Introduction}

High-definition (HD) semantic maps are an essential module for autonomous driving. Traditional pipelines to construct such HD semantic maps involve capturing point clouds beforehand, building globally-consistent maps using SLAM, and annotating semantics in the maps. This paradigm, though producing accurate HD maps and adopted by many autonomous driving companies, requires a vast amount of human efforts. 

As an alternative, we investigate scalable and affordable autonomous driving solutions, \eg minimizing human efforts in annotating and maintaining HD maps.
%Our ultimate goal is to build an autonomous driving system that is solely based on local observations. 
To that end, we introduce a novel {semantic map learning} framework that makes use of on-board sensors and computation to estimate vectorized local semantic maps. 
% Our framework does not maintain or update global HD maps, removing humans' necessity in the loop, {making it more scalable than the previous pipeline. 
% Though the proposed framework currently does not produce map elements as accurate, it offers system developers another set of trade-off choice between scalability and accuracy.
Of note, our framework \textit{does not} aim to replace global HD map reconstruction, instead to provide a simple way to predict local semantic maps for real-time motion prediction and planning. 

\begin{figure*}[!t]
\centering
\includegraphics[width=0.8\textwidth]{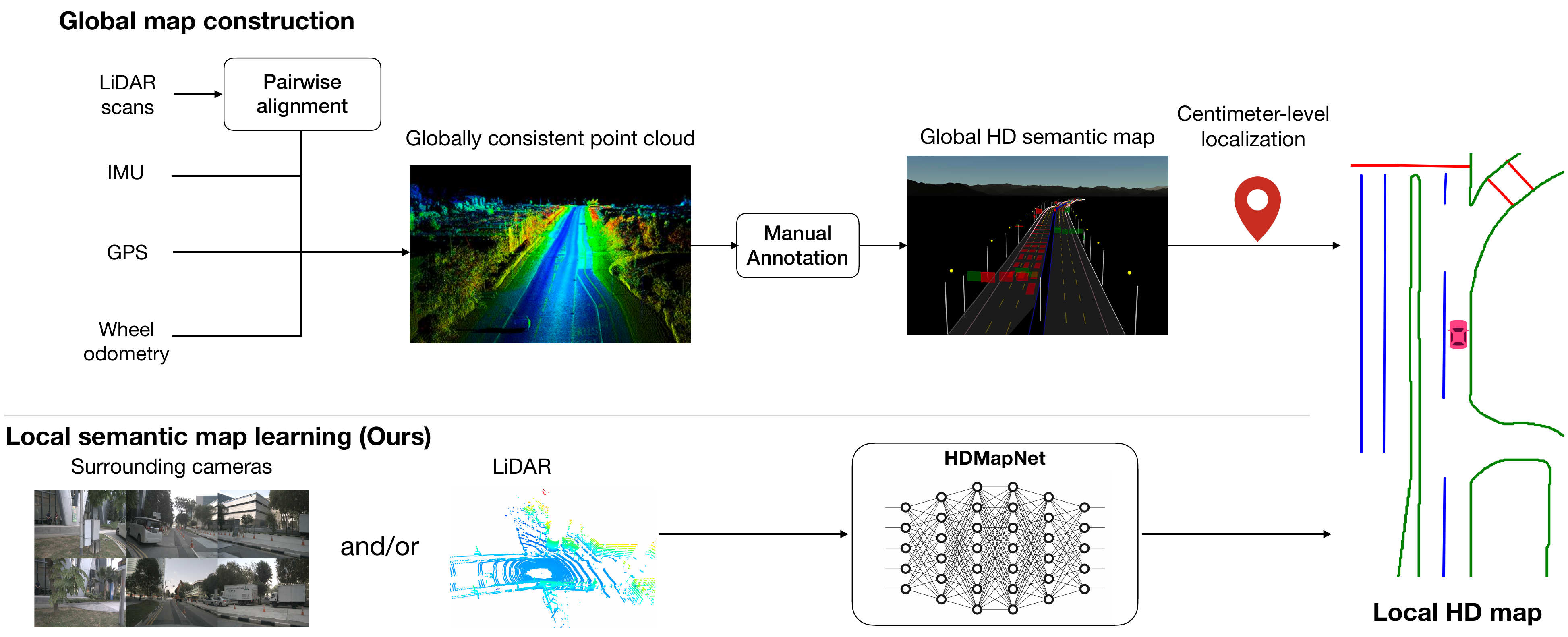}
\captionof{figure}{
{
In contrast to pre-annotating global semantic maps, we introduce a novel local map learning framework that makes use of on-board sensors to estimate local semantic maps. \label{fig:teaser}
\vspace{-5mm}
}
}
\end{figure*}

We propose a semantic map learning method named HDMapNet, which produces vectorized map elements from images of the surrounding cameras and/or from point clouds like LiDARs.
We study how to effectively transform perspective image features to bird's-eye view features when depth is missing. We put forward a novel view transformer that consists of both neural feature transformation and geometric projection. 
% uses neural networks to transform features from perspective view to birds-eye view in the camera coordinate
Moreover, we investigate whether point clouds and camera images complement each other in this task. We find different map elements are not equally recognizable in a single modality. To take the best from both worlds, our best model combines point cloud representations with image representations. This model outperforms its single-modal counterparts by a significant margin in all categories. To demonstrate the practical value of our method, we generate a locally-consistent map using our model in Figure~\ref{fig:temporal}; the map is immediately applicable to real-time motion planning.  
% Our model consists of three major modules: an image encoder that encodes perspective view images and transforms the features into the birds-eye view; a 3D encoder that embeds point clouds into birds-eye view feature space; a BEV decoder that takes birds-eye features and predicts semantic labels and instance IDs for each birds-eye view position.
% a perspective view network to encode camera inputs to feature space; a novel projection module to convert the perspective features to birds-eye view features; a birds-eye view network to predict semantics and align map elements. 
% Unlike traditional methods, our pipeline computes the online HD maps based on local camera observations.

Finally, we propose comprehensive ways to evaluate the performance of map learning. These metrics include both semantic level and instance level evaluations as map elements are typically represented as object instances in HD maps. 
On the public NuScenes dataset, HDMapNet improves over existing methods by 12.1 IoU on semantic segmentation and 13.1 mAP on instance detection. 

%Also, these metrics measure the performance from both \textit{Eulerian} and \textit{Lagrangian} perspectives. 
%Our method takes 360\degree images as input and predict HD map elements with a perspective view network and a birds-eye view network.
% we argue that you need to build real-time local map
%we need a figure 1 shows how traditional HDmap is built and how our hdmap is built 
% In this paper, instead of maintaining the freshness of highly accurate and richly annotated HD maps, and using centimeter level localization, to build the local maps of the scenes surrounding autonomous vehicles. We advocate for real-time onboard sensing of local HD map. Traditional HD maps, together with high precision localization, help autonomous vehicle perceive the static environment, such as lane markings, road markings, dividers and road boundary around the vehicle. Given recent advance in autonomous driving sensors, such information can be estimated through surrounding camera

To summarize, our contributions include the following:
\begin{itemize} 
\setlength\itemsep{0em}
\item We propose a novel online framework to construct HD semantic maps from the sensory observations, and together with a method named HDMapNet. % Our model improves over existing methods by 12.1 IoU on semantic segmentation and 13.1 mAP on instance detection. 
\item We come up with a novel feature projection module from perspective view to bird's-eye view. This module models 3D environments implicitly and considers the camera extrinsic explicitly. 
% \item We provide in-depth analysis and visualizations to understand our models.
\item We develop comprehensive evaluation protocols and metrics to facilitate future research.
\end{itemize}

% \vspace{-3mm}
% \vspace{-2mm}
\section{Related Work}
% \vspace{-3mm}

{\noindent\textbf{Semantic map construction.}}
Most existing {HD semantic maps} are annotated either manually or semi-automatically on LiDAR point clouds of the environment, merged from LiDAR scans collected from survey vehicles with high-end GPS and IMU. SLAM algorithms are the most commonly used algorithms to fuse LiDAR scans into a highly accurate and consistent point cloud. First, pairwise alignment algorithm like ICP~\cite{besl1992method}, NDT~\cite{biber2003normal} and their variants~\cite{segal2009generalized} are employed to match LiDAR data at two nearby timestamps using semantic~\cite{yu2015semantic} or geometry information~\cite{pomerleau2015review}. Second, estimating accurate poses of ego vehicle is formulated as a non-linear least-square problem~\cite{lawson1995solving} or a factor graph~\cite{dellaert2012factor} which is critical to build a globally consistent map. Yang~\etal~\cite{yang2018robust} presented a method for reconstructing maps at city scale based on the pose graph optimization under the constraint of pairwise alignment factor. To reduce the cost of manual annotation of semantic maps, \cite{jiao2018machine,Mi_2021_CVPR} proposed several machine learning techniques to extract static elements from fused LiDAR point clouds and cameras. However, it is still laborious and costly to maintain an HD semantic map since it requires high precision and timely update. In this paper, we argue that our proposed local semantic map learning task is a potentially more scalable solution for autonomous driving.

%\textbf{Road extraction from overhead imagery.}
%In remote sensing, extracting roads from satellite and aerial images has been an active line of research~\cite{richards_2013, wegner2013higher, wegner2015road}.
% Wegner~\etal~\cite{wegner2013higher, wegner2015road} formulated CRF models to extract road networks. 
%Mnih~\etal~\cite{mnih2010learning,mnih2012learning} were the first to extract roads using neural networks by formulating the task as segmentation, and many works followed this formulation~\cite{mohan2014deep, marmanis2016semantic, marmanis2018classification}.
%Later works~\cite{Mattyus_2017_ICCV,ventura2018iterative,bastani2018roadtracer} applied image segmentation on aerial images and then perform post graph optimization.
%More recently, Liang~\etal~\cite{liang2019convolutional} and Homayounfar~\etal~\cite{Homayounfar_2019_ICCV} proposed convolutional recurrent networks to replace post-processing steps for road extraction. 
%Unlike this line of work, we aim to extract map elements from perspective view images observed by a moving vehicle.

\noindent\textbf{Perspective view lane detection.}
The traditional perspective-view-based lane detection pipeline involves local image feature extraction (\eg color, directional filters~\cite{chiu2005lane,loose2009kalman,zhou2010novel}) , line fitting (\eg Hough transform~\cite{illingworth1988survey}), image-to-world projection, \etc
With the advances of deep learning based image segmentation and detection techniques~\cite{alvarez2012road,zhou2017scene,ess2009segmentation,cityscape}, researchers have explored more data-driven approaches. Deep models were developed for road segmentation~\cite{yu2018bdd100k,neuhold2017mapillary}, lane detection~\cite{wang2018lanenet,neven2018towards}, drivable area analysis~\cite{liu2018segmentation}, \etc
% These approaches all output pixel-level results in the image coordinates.
More recently, models were built to give 3D outputs rather than 2D. Bai~\etal~\cite{bai2018} incorporated LiDAR signals so that image pixels can be projected onto the ground. Garnett~\etal~\cite{garnett20193d} and Guo~\etal~\cite{guo2020gen} used synthetic lane datasets to perform supervised training on the prediction of camera height and pitch, so that the output lanes sit in a 3D ground plane.
Beyond detecting lanes, our work outputs a consistent {local semantic map} around the vehicle from surround cameras or LiDARs.

\noindent\textbf{Cross-view learning.}
Recently, some efforts have been made to study cross-view learning to facilitate robots' surrounding sensing capability. 
% These works share a similar framework that encodes image features in perspective view and transforms features into bird's-eye view to be decoded into various sensing targets.
%However, the models they use to transform into bird's-eye views differ.
Pan~\cite{Pan_2020} used MLPs to learn the relationship between perspective-view feature maps and bird's-eye view feature maps. 
Roddick and Cipolla ~\cite{roddick2020predicting} applied 1D convolution on the image fetures along the horizontal axis to predict bird's-eye view.
Philion and Fidler~\cite{philion2020lift} predicted the depth of monocular cameras and project image features into bird's-eye view using soft attention.
Our work focuses on the crucial task of local semantic map construction that we use cross-view sensing methods to generate map elements in a vectorized form. %We show that our model outperforms prior projection methods in Section~\ref{sec:experiment}.
Moreover, our model can be easily fused with LiDAR input to further improve its accuracy.

% \vspace{-4mm}
\section{Semantic Map Learning}
% \vspace{-4mm}
We propose semantic map learning, a novel framework that produces {local high-definition semantic maps}. It takes sensor inputs like camera images and LiDAR point clouds, and outputs vectorized map elements, such as lane dividers, lane boundaries and pedestrian crossings.
We use $\I$ and $P$ to denote the images and point clouds, respectively. Optionally, the framework can be extended to include other sensor signals like radars. We define $\M$ as the map elements to predict.

% In Section~\ref{sec:hdmapnet}, we will discuss our proposed method for online HD map learning; in Section~\ref{sec:evaluation}, we will develop metrics to evaluate map learning methods.

% \vspace{-3mm}
\subsection{HDMapNet}
% \vspace{-3mm}
\label{sec:hdmapnet}

\begin{figure*}[!t]
\centering
\includegraphics[width=0.9\textwidth]{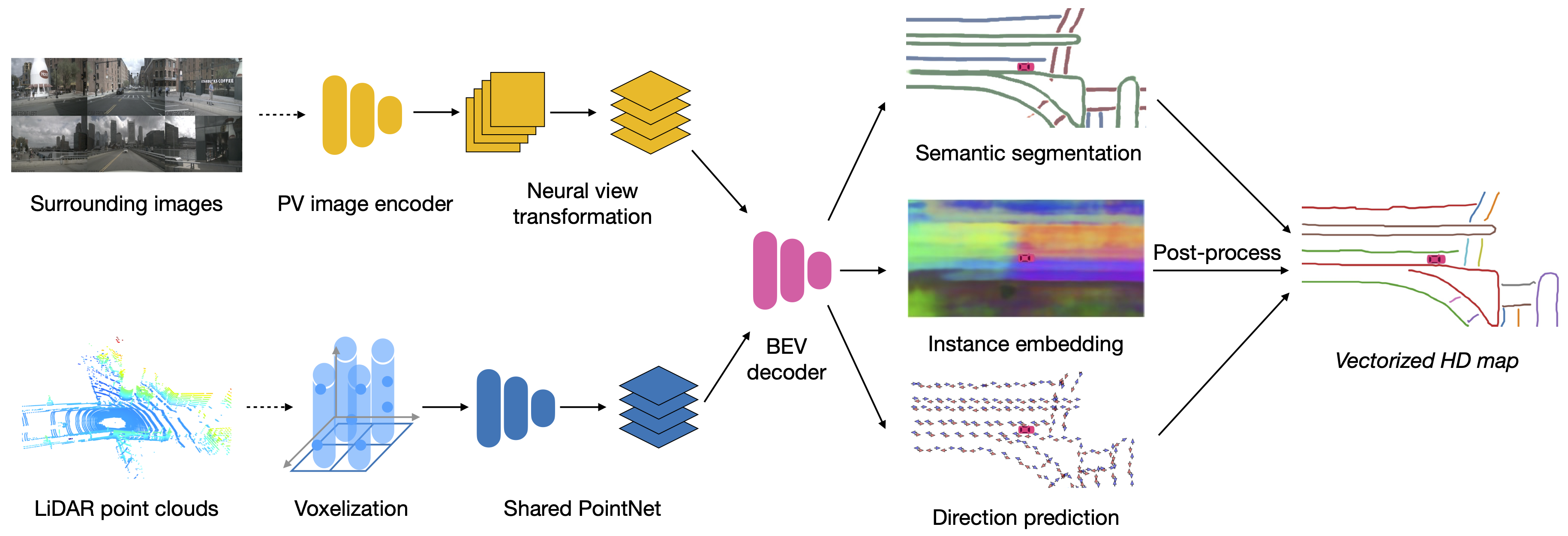}
\caption{Model overview. HDMapNet works with either or both of images and point clouds, outputs semantic segmentation, instance embedding and directions, and finally produces a vectorized local semantic map. Top left: image branch. Bottom left: point cloud branch. Right: HD semantic map. \label{fig:overview}.
\vspace{-5mm}
}
\end{figure*}

% In this section, we detail our approach to the HD map learning framework.
% % We formulate the problem in \S\ref{sec:formulation}. 
% First, we present the proposed model HDMapNet in \S\ref{sec:hdmapnet}. We discuss three critical components of our model: a perspective view image encoder and a view transformer \S\ref{sec:image}, a point cloud encoder~\S\ref{sec:pointpillar}, and a bird's-eye view~\S\ref{sec:decoder}. Finally, we discuss the evaluation metrics in~\S\ref{sec:evaluation}. We propose both semantic level metrics (\S\ref{sec:semantic_metric}) and instance-level metrics (\S\ref{sec:instance_metric})
%\vspace{-0.5mm}

% (see Figure~\ref{fig:teaser}) 
%Classic algorithms typically deploy complicated pipelines to merge data from multiple sensors and require significant human efforts in the loop. 
Our {semantic map learning} model, named HDMapNet, predicts map elements $\M$ from single frame $\I$ and $P$ with neural networks directly.
An overview is shown in Figure~\ref{fig:overview},
four neural networks parameterize our model: a perspective view image encoder $\phi_\I$ and a neural view transformer $\phi_\V$ in the image branch, a pillar-based point cloud encoder $\phi_P$, and a map element decoder $\phi_\M$. We denote our HDMapNet family as HDMapNet(Surr), HDMapNet(LiDAR), HDMapNet(Fusion) if the model takes only surrounding images, only LiDAR, or both of them as input.

% \vspace{-2mm}
\subsubsection{Image encoder} Our image encoder has two components, namely perspective view image encoder and neural view transformer.

% \vspace{-1mm}
\label{sec:image}
\noindent\textbf{Perspective view image encoder.} Our image branch takes perspective view inputs from $N_m$ surrounding cameras, covering the panorama of the scene. Each image $\I_i$ is embedded by a shared neural network $\phi_\I$ to get perspective view feature map  $\F_{\I_i}^{\mathrm{pv}} \subseteq  \R^{H_{\mathrm{pv}}\times W_{\mathrm{pv}} \times K}$
% \begin{equation}
% \centering
% \F_{\I_i}^{\mathrm{pv}} = \phi_\I (\I_i),
% \end{equation}
% where $\F_{\I_i}^{\mathrm{pv}} \subseteq  \R^{H_{\mathrm{pv}}\times W_{\mathrm{pv}} \times K}$ is the feature map for each image in the perspective view; 
where $H_{\mathrm{pv}}$, $W_{\mathrm{pv}}$, and $K$ are the height, width, and feature dimension respectively. 

\begin{figure}[!t]
\centering 
\includegraphics[width=1.0\columnwidth]{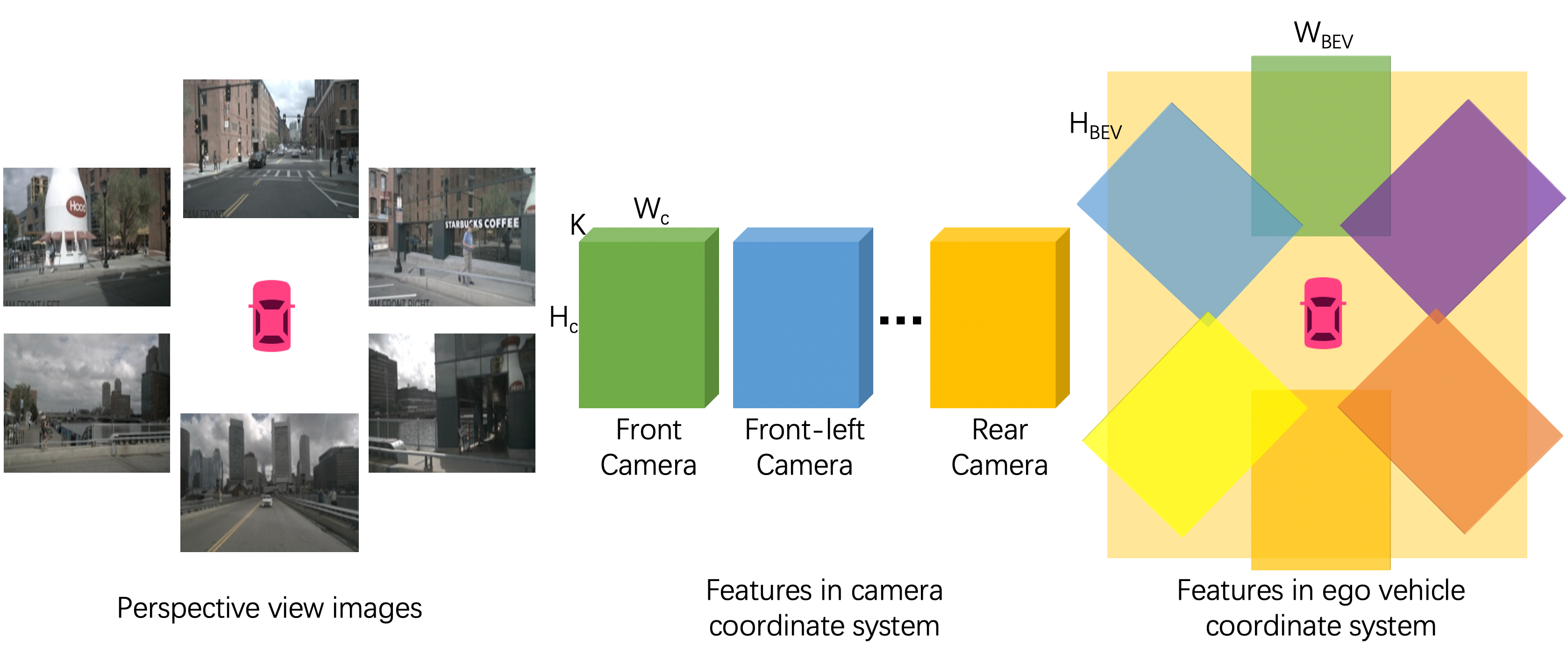}
\caption{Feature transformation. Left: the 6 input images in the perspective view. Middle: 6 feature maps in camera coordinate system, which are obtained by extracting features using image encoder and transforming these features with MLP; each feature map (in different colors) covers a certain area. Right: the feature map (in orange) in the ego vehicle coordinate system; this is fused from 6 feature maps and transformed to the ego vehicle coordinate system with camera extrinsics. \label{fig:nvt}}
\vspace{-5mm}
\end{figure}

\noindent\textbf{Neural view transformer.} As shown in Figure~\ref{fig:nvt}, we first transform image features from perspective view to camera coordinate system and then to bird's-eye view.
% which utilize both neural networks and geometric information.
% Each image feature map is warped into the camera coordinate system 
The relation of any two pixels between perspective view and camera coodinate system is modeled by a multi-layer perceptron $\phi_{\V_i}$:
\begin{equation}
\centering
\F_{\I_i}^c[h][w] = \phi_{\V_i}^{hw}(\F_{\I_i}^{\mathrm{pv}}[1][1], \dots, \F_{\I_i}^{\mathrm{pv}}[H_{\mathrm{pv}}][W_{\mathrm{pv}}])
% \F_{\I_i}^c = \phi_\V (\F_{\I_i}),
\end{equation}
where $\phi_{\V_i}^{hw}$ models the relation between feature vector at position $(h, w)$ in the camera coodinate system and every pixel on the perspective view feature map. We denote $H_c$ and $W_c$ as the top-down spatial dimensions of $F_{\I}^c$.
% $\F_{\I_i}^c \subseteq  \R^{H_c\times W_c \times K}$ and $H_c$ and $W_c$ are the top-down spatial dimensions in the camera coordinate system. 
The bird's-eye view (ego coordinate system) features $\F_{\I_i}^{\mathrm{bev}} \subseteq  \R^{H_{\mathrm{bev}}\times W_{\mathrm{bev}} \times K}$ is obtained by transforming the features $\F_{\I_i}^c$ using geometric projection with camera extrinsics, where $H_{\mathrm{bev}}$ and $W_{\mathrm{bev}}$ are the height and width in the bird's-eye view. The final image feature $\F_\I^{\mathrm{bev}}$ is an average of $N_m$ camera features.

% \vspace{-2mm}
\subsubsection{Point cloud encoder}
% \vspace{-1mm}
\label{sec:pointpillar}
% \begin{figure}
% \includegraphics[width=\columnwidth]{section/figs/pp}
% \caption{neural view transformation\label{fig:pp}}
% \end{figure}
Our point cloud encoder $\phi_P$ is a variant of PointPillar~\cite{Lang_2019_CVPR} with dynamic voxelization~\cite{Zhou2019EndtoEndMF}, which divide the 3d space into multiple pillars and learn feature maps from pillar-wise features of pillar-wise point clouds. The input is $N$ lidar points in the point cloud. For each point $p$, it has three-dimensional coordinates and additional $K$-dimensional features represented as $f_p \subseteq \R^{K+3}$. 
%a three-dimensional point cloud with $N$ points $P = \{p_i\}_{i=0}^{N-1} \subseteq \R^3$ with $K$-dimensional features $\F_{P} = \{f_i\}_{i=0}^{N-1} \subseteq \R^K$. 
%We define functions $F_V(p_i)$ that returns the index $j$ of $p_i$'s corresponding pillar $v_j$. 

When projecting features from points to bird's-eye view, multiple points can potentially fall into the same pillar. We define $P_j$ as the set of points corresponding to pillar $j$. To aggregate features from points in a pillar, a PointNet~\cite{Qi_2017_CVPR} (denoted as $\mathrm{PN}$) is warranted, where 
\vspace{-2mm}
\begin{equation}\label{eq:pillarfeature}
     \begin{split}
        f^{\mathrm{pillar}}_j = \mathrm{PN}(\{f_p | \forall p \in P_j\}).
     \end{split}
\end{equation}
%\vspace{-1mm}
Then, pillar-wise features are further encoded through a convolutional neural network $\phi_{\mathrm{pillar}}$. We denote the feature map in the bird's-eye view as $\F^{\mathrm{bev}}_P$. 
% and the feature for pixel j as $f^{\mathrm{bev}}_j$.

% \vspace{-2mm}
\subsubsection{Bird's-eye view decoder}
% \vspace{-1mm}
\label{sec:decoder}
The map is a complex graph network that includes instance-level and directional information of lane dividers and lane boundaries. Instead of pixel-level representation, lane lines need to be vectorized so that they can be followed by self-driving vehicles. Therefore, our BEV decoder $\phi_\M$ not only outputs semantic segmentation but also predicts instance embedding and lane direction. A post-processing process is applied to cluster instances from embeddings and vectorize them.

\noindent\textbf{Overall architecture.} The BEV decoder is a fully convolutional network (FCN)~\cite{shelhamer2017fully} with 3 branches, namely semantic segmentation branch, instance embedding branch, and direction prediction branch. The input of BEV decoder is image feature map $F_{\I}^{\mathrm{bev}}$ and/or point cloud feature map $F_{P}^{\mathrm{bev}}$, and we concatenate them if both exist.

\noindent\textbf{Semantic prediction.} The semantic prediction module is a fully convolutional network (FCN)~\cite{shelhamer2017fully} %$\phi_{\mathrm{semantic}}$. The bird's-eye view feature map $\F^{\mathrm{bev}}$ is fed into this FCN decoder, and per-pixel semantic predictions are made. 
We use cross-entropy loss for the semantic prediction.

\noindent\textbf{Instance embedding.} Our instance embedding module seeks to cluster each bird's-eye view embedding. For ease of notation, we follow the exact definition in~\cite{Brabandere_2017_CVPR_Workshops}:
$C$ is the number of clusters in the ground truth, $N_c$ is the number of elements in cluster $c$, $\mu_c$ is the mean embedding of cluster c, $\|\cdot\|$ is the L2 norm, and $[x]_+$ = max(0, x) denotes the element maximum. $\delta_v$ and $\delta_d$ are respectively the margins for the variance and distance loss. 
% First, the instance embedding module uses a FCN $\phi_{\mathrm{instance}}$ to encode each bird's-eye feature $f^{bev}_{j}$ to $f^{\mathrm{instance}}_{j}$. Then, 
The clustering loss $L$ is computed by:
\begin{align}
     &L_{var} = \frac{1}{C} \sum^C_{c=1} \frac{1}{N_c} \sum^{N_c}_{j=1} [\|\mu_c - f^{\mathrm{instance}}_{j}\| - \delta_v]_+^2 \;\;\;\; \\
     &L_{dist} = \frac{1}{C(C-1)} \sum_{c_A \neq c_B \in C}[2 \delta_d - \|\mu_{c_A} - \mu_{c_B}\|]_+^2, \\
     &L = \alpha L_{var} + \beta L_{dist} \label{eq:discriminative_loss}.
\end{align}

% \begin{equation}
%     L_{var} = \frac{1}{C} \sum^C_{c=1} \frac{1}{N_c} \sum^{N_c}_{j=1} [\|\mu_c - f^{\mathrm{instance}}_{j}\| - \delta_v]_+^2 \;\;\;\; \\
% \end{equation}

% \begin{equation}
%     L_{dist} = \frac{1}{C(C-1)} \sum_{c_A \neq c_B \in C}[2 \delta_d - \|\mu_{c_A} - \mu_{c_B}\|]_+^2,
% \end{equation}
% %\end{equation}
% % \begin{equation}
% %     L_{reg} = \frac{1}{C} \sum^C_{c=1} \|\mu_c\|,
% % \end{equation}
% \begin{equation}
%     \label{eq:discriminative_loss}
%     % L = \alpha L_{var} + \beta L_{dist} + \gamma L_{reg}.
%     L = \alpha L_{var} + \beta L_{dist}.
% \end{equation}

% We use the default $\alpha, \beta, $ and $\gamma$ as in~\cite{Brabandere_2017_CVPR_Workshops}.

\noindent\textbf{Direction prediction.}
{
Our direction module aims to predict directions of lanes from each pixel $C$. The directions are discretized into $N_d$ classes uniformed distributed on a unit circle.}
% The direction is discretized into 36 classes (0, 10, ..., 350 degrees), with an extra label indicating ``no direction". 
% There are three types of pixel: the pixel outside the lane, which has no direction; the pixel on the edge of the lane, which has one direction; and the pixel inside the lane, which has two direction. We don't consider the pixel outside the lane (no backpropagation during training) and for the the edge pixel, we give it two labels: the actual direction and "no direction".
By classifying direction $D$ of current pixel $C_{now}$, the next pixel of lane $C_{next}$ can be obtained as $C_{next} = C_{now} + \Delta_{step} \cdot D$, where $\Delta_{step}$ is a predefined step size. Since we don't know the direction of the lane, we cannot identify the forward and backward direction of each node. Instead, we treat both of them as positive labels. Concretely, the direction label of each lane node is a $N_d$ vector with 2 indices labeled as 1 and others labeled as 0.
% Since each node on the lane has two directions, the direction estimation problem is formulated as a multi-class two-label classification problem, where the top-two indices are used as the predicted directions. 
Note that most of the pixels on the topdown map don't lie on the lanes, which means they don't have directions. The direction vector of those pixels is a zero vector and we never do backpropagation for those pixels during training. We use softmax as the activation function for classification.

\begin{figure*}[t!]
    \centering
    \begin{subfigure}[t]{.3\textwidth}
        \centering
        \includegraphics[width=0.8\linewidth]{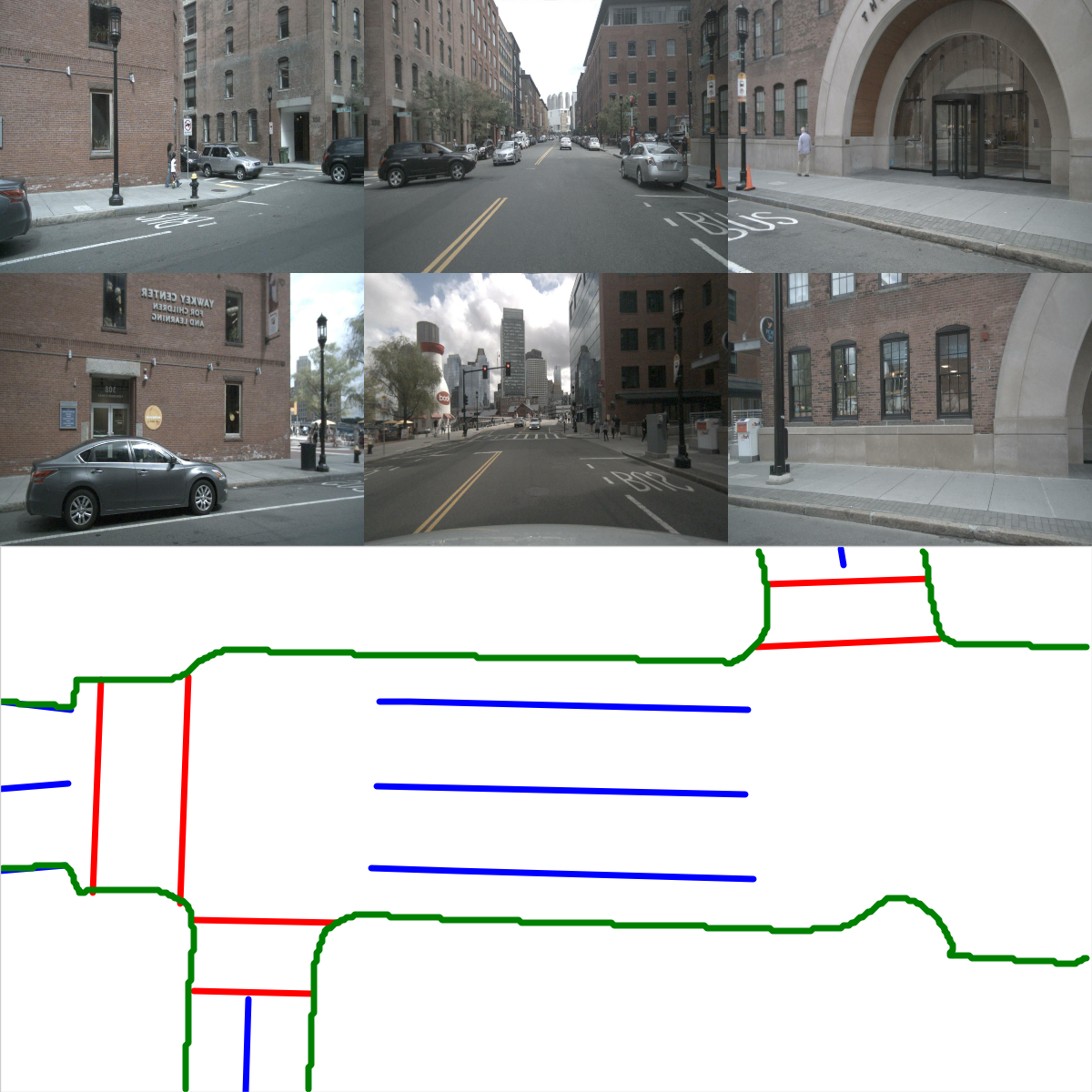}
        % \vspace{-0.5em}
        \caption*{Ground Truth}
    \end{subfigure}
    \begin{subfigure}[t]{.3\textwidth}
        \centering
        \includegraphics[width=0.8\linewidth]{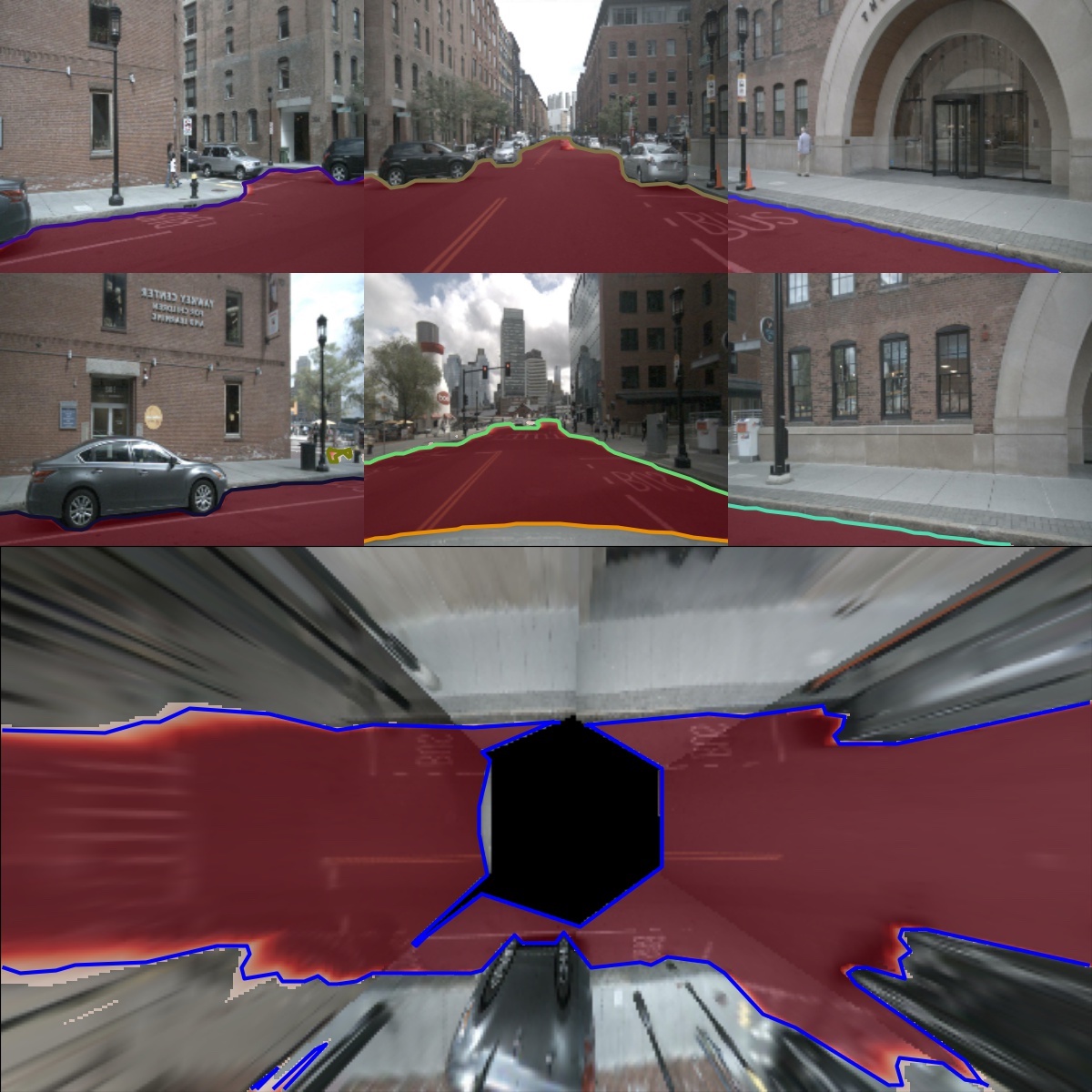}
        % \vspace{-0.5em}
        \caption*{IPM}
    \end{subfigure}
    \begin{subfigure}[t]{.3\textwidth}
        \centering
        \includegraphics[width=0.8\linewidth]{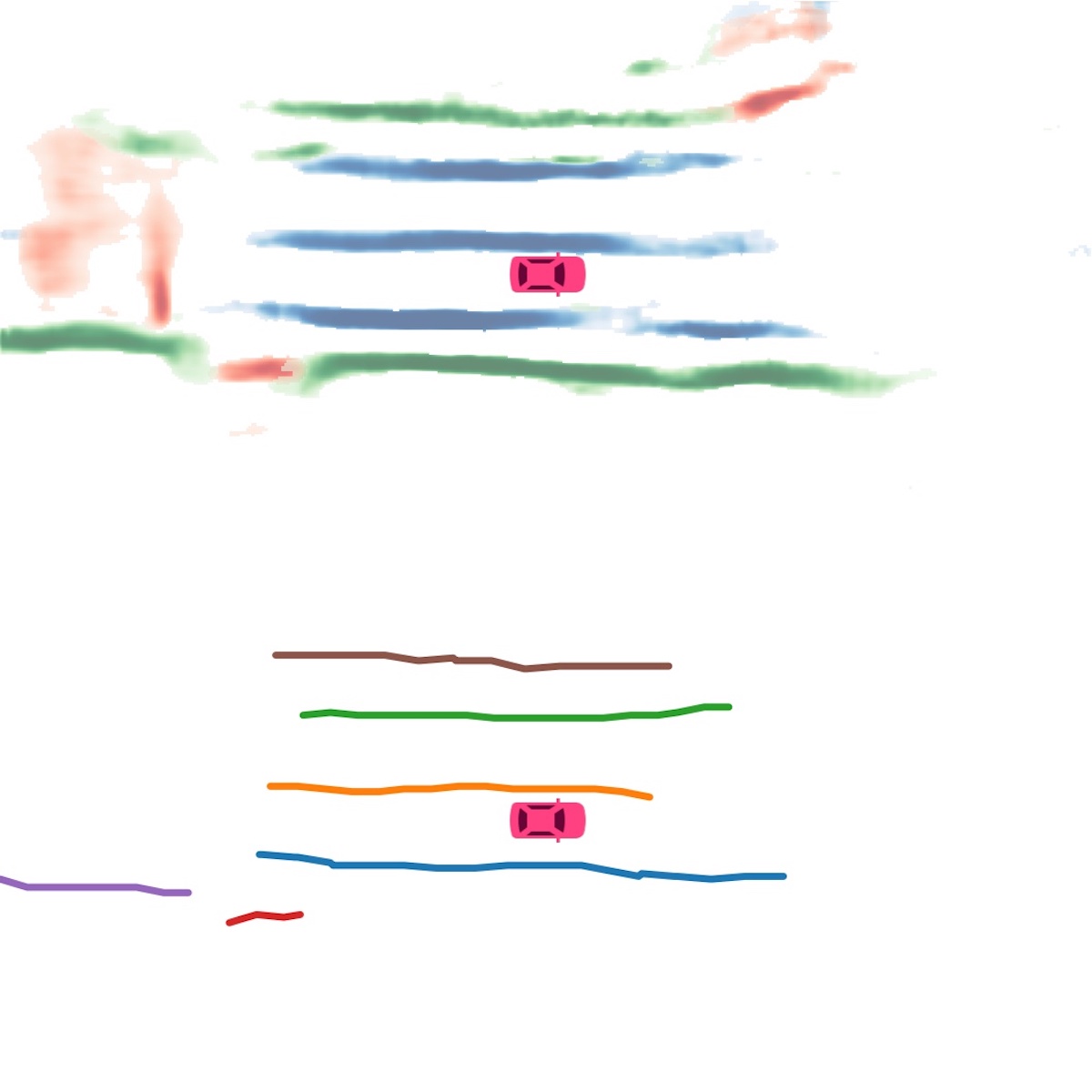}
        % \vspace{-0.5em}
        \caption*{IPM(B)}
    \end{subfigure}
    
    % \vspace{1em}
    
    \begin{subfigure}[t]{.3\textwidth}
        \centering
        \includegraphics[width=0.8\linewidth]{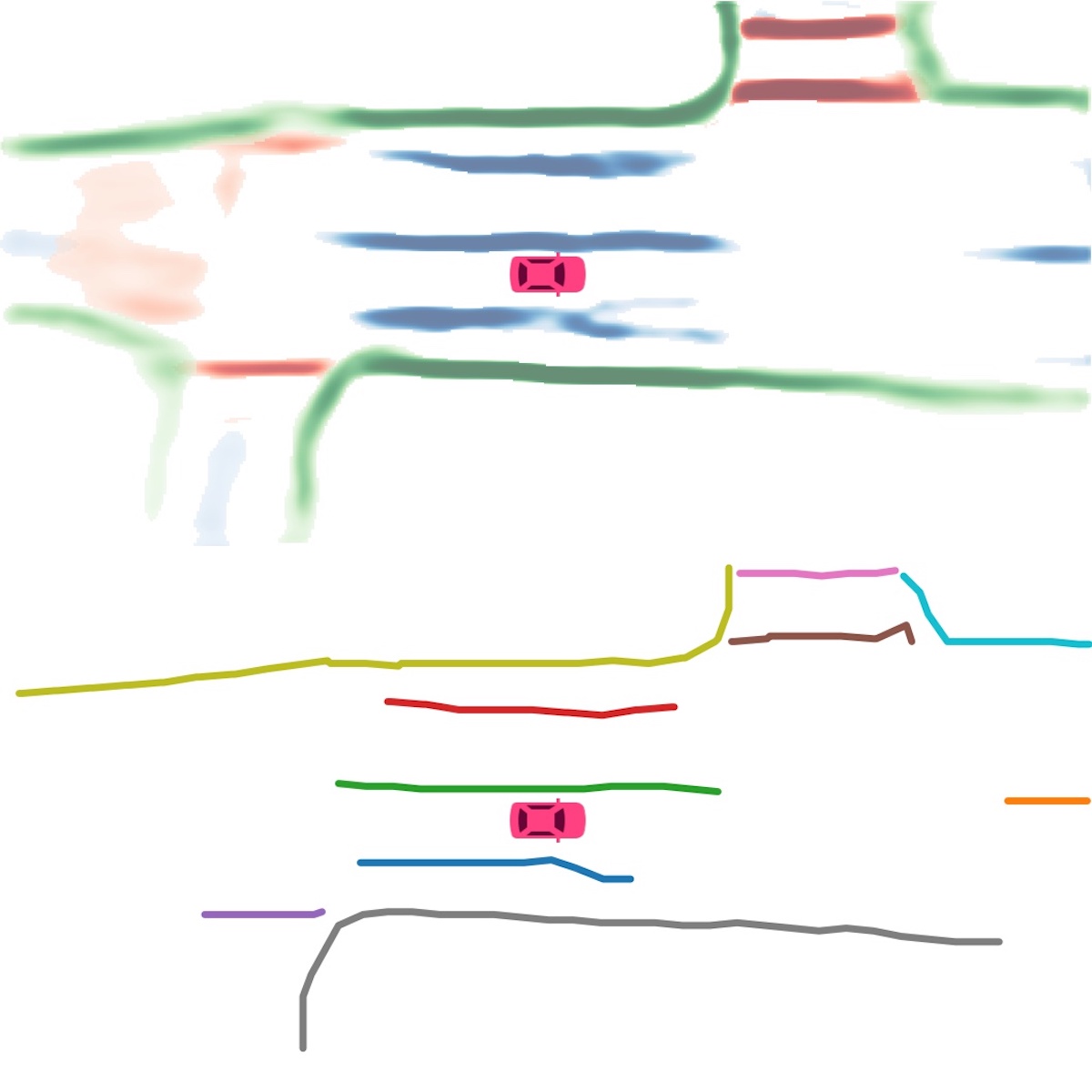}
        % \vspace{-0.5em}
        \caption*{IPM(CB)}
    \end{subfigure}
    \begin{subfigure}[t]{.3\textwidth}
        \centering
        \includegraphics[width=0.8\linewidth]{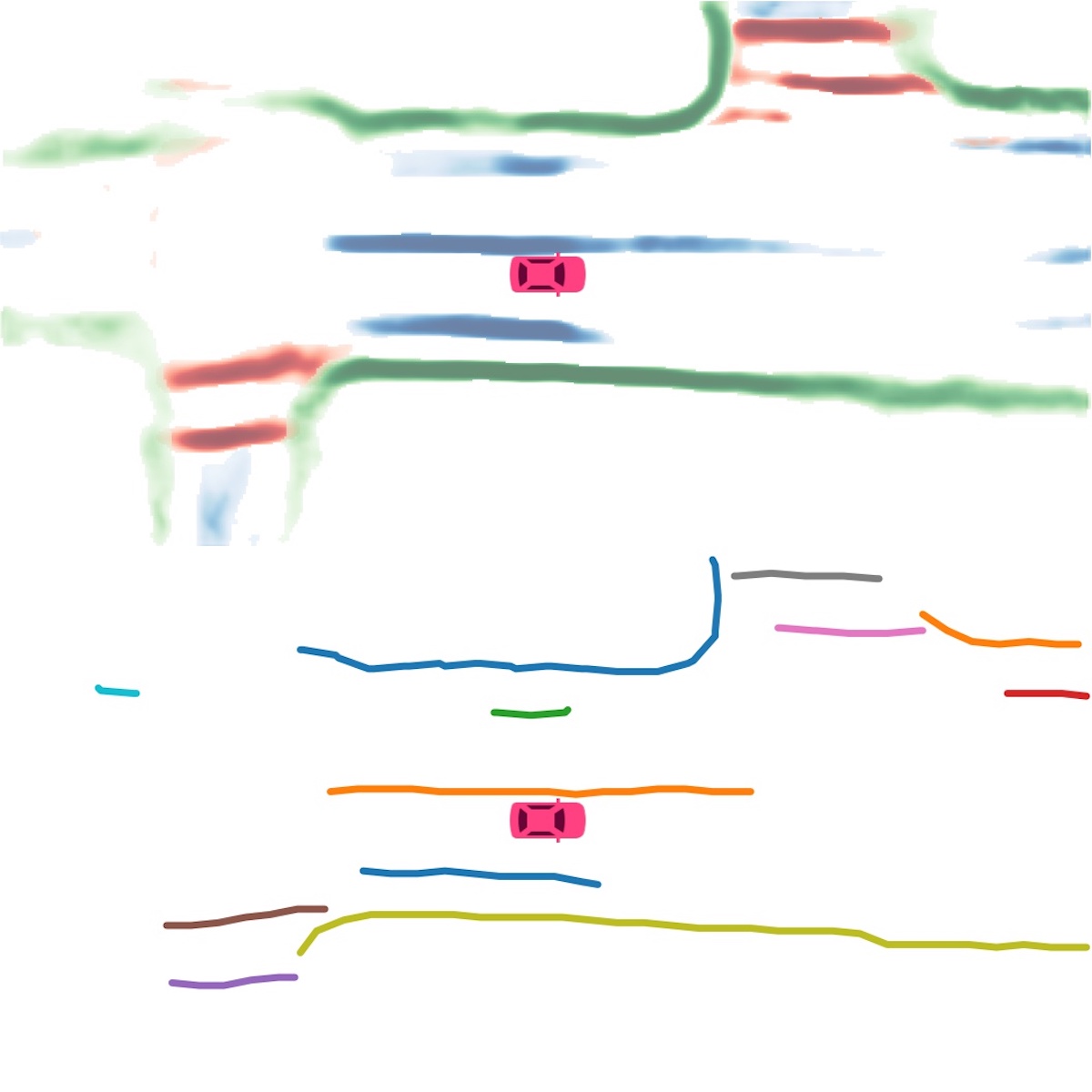}
        % \vspace{-0.5em}
        \caption*{Lift-Splat-Shoot}
    \end{subfigure}
    \begin{subfigure}[t]{.3\textwidth}
        \centering
        \includegraphics[width=0.8\linewidth]{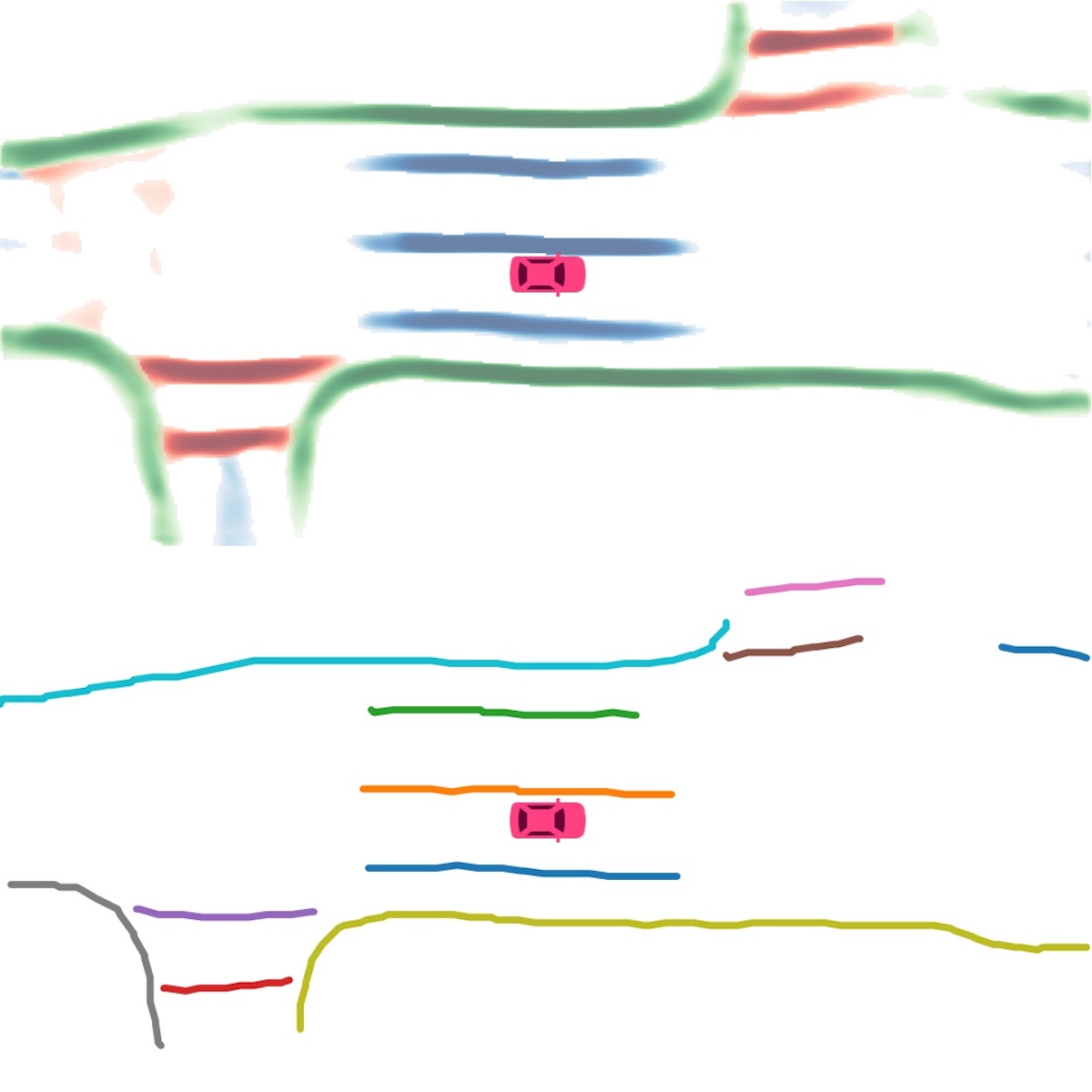}
        % \vspace{-0.5em}
        \caption*{VPN}
    \end{subfigure}

    % \vspace{1em}
    
    \begin{subfigure}[t]{.3\textwidth}
        \centering
        \includegraphics[width=0.8\linewidth]{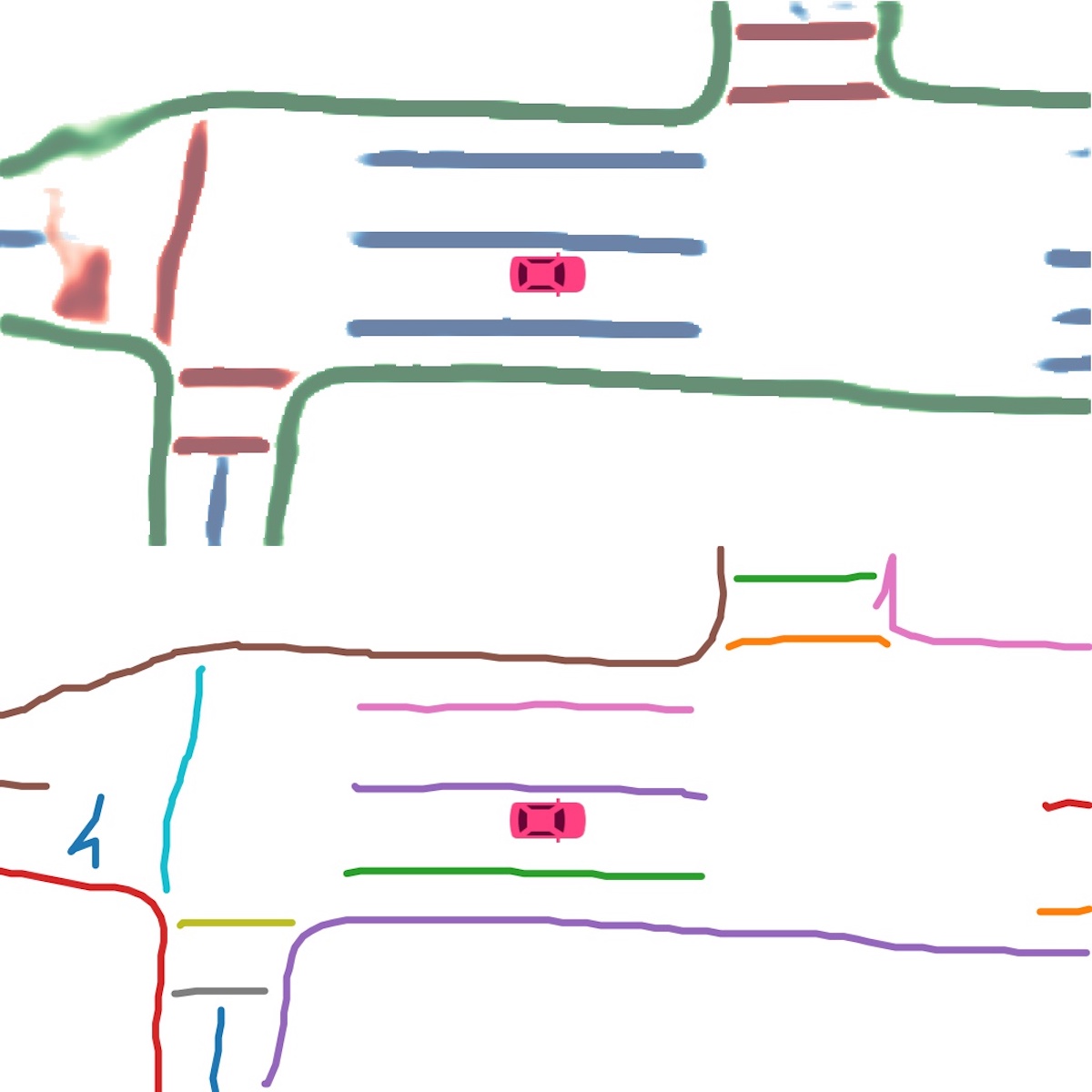}
        % \vspace{-0.5em}
        \caption*{HDMapNet(Surr)}
    \end{subfigure}
    \begin{subfigure}[t]{.3\textwidth}
        \centering
        \includegraphics[width=0.8\linewidth]{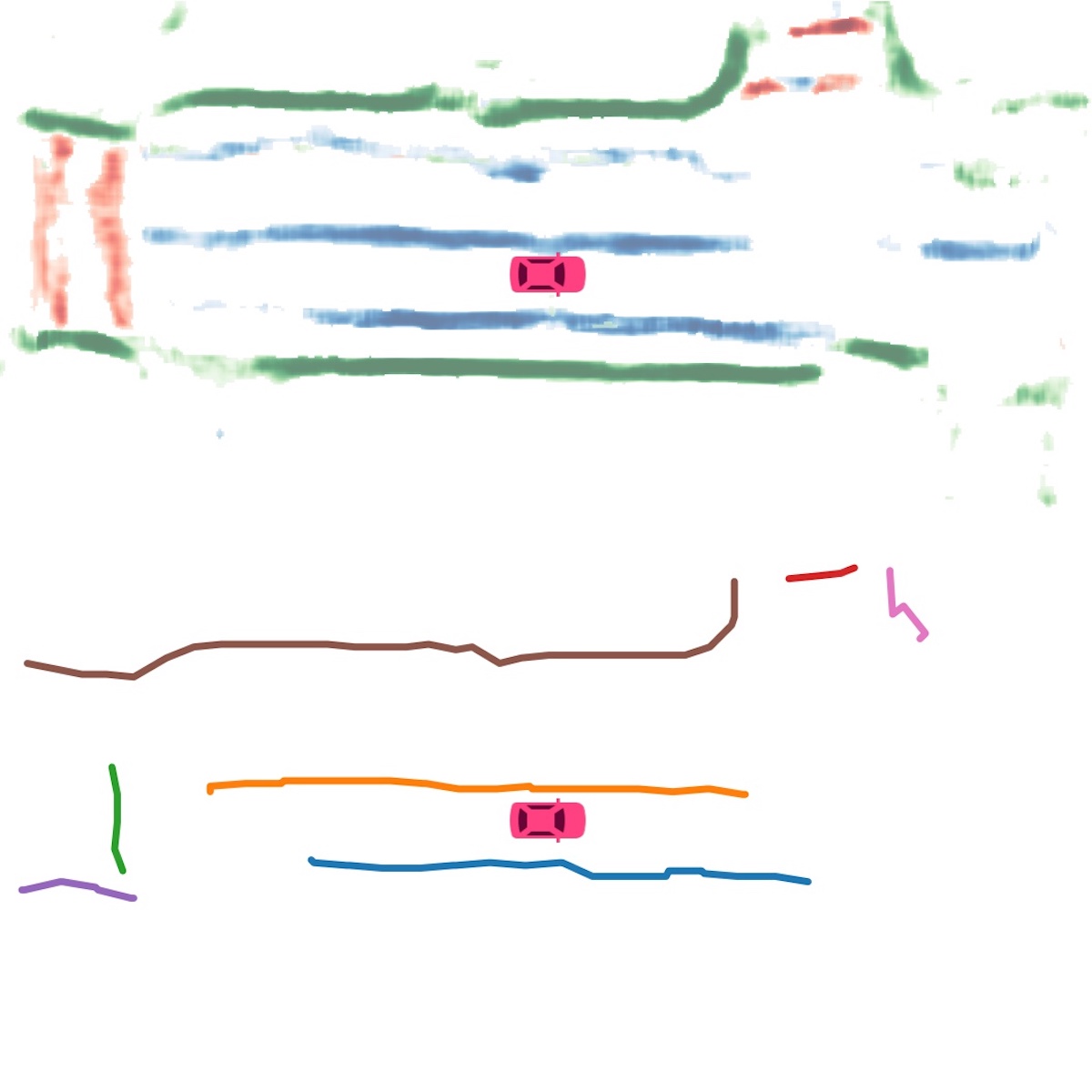}
        % \vspace{-0.5em}
        \caption*{HDMapNet(LiDAR)}
    \end{subfigure}
    \begin{subfigure}[t]{.3\textwidth}
        \centering
        \includegraphics[width=0.8\linewidth]{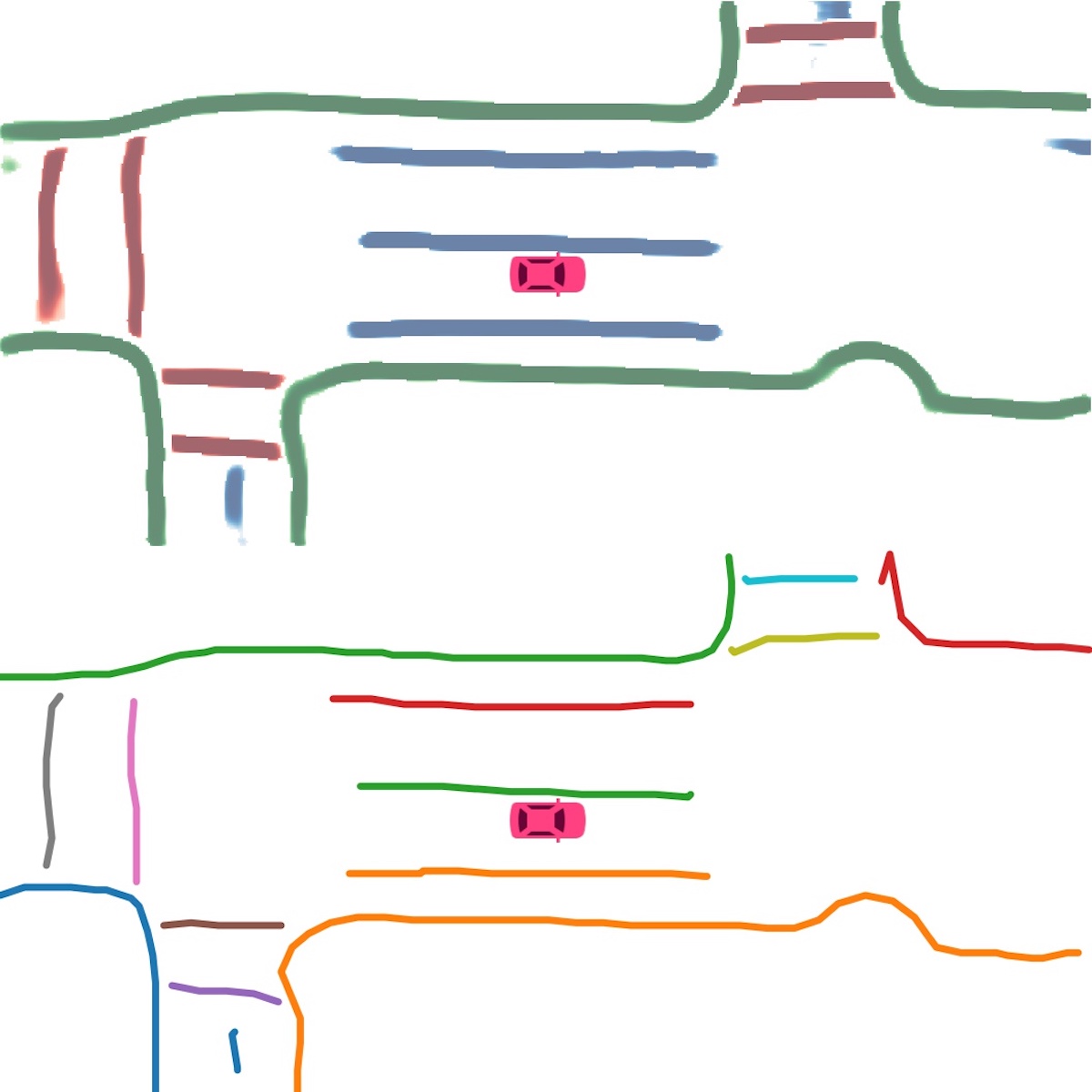}
        % \vspace{-2.5em}
        \caption*{HDMapNet(Fusion)}
    \end{subfigure}

    \caption{Qualitative results on the validation set. Top left: the surrounding images and the ground-truth local semantic map annotations. IPM: the lane segmentation result in the perspective view and the bird's-eye view. % We can see the distortion of lanes after the IPM, and the same lane is identified as different on the perspective view. 
    Others: the semantic segmentation results and the vectorized instance detection results. 
    %Dashed circles: curves are predicted accurately in our HDMapNet but not in other methods. 
    %It can be seen that camera based method is good at predicting straight lanes, despite the obstruction of cars. On the other hand, only LiDAR based method detect the curvature at the bottom-left corner, which is very subtle. This indicate the promise to combine multiple sensors to construct a precise local HD Map. Note that Our HDMapNet produce a much more detailed and precise local HD Map than all other methods with high confidence.
    }
    \vspace{-3mm}
    \label{fig:results}
\end{figure*}

\noindent\textbf{Vectorization.} During inference, we first cluster instance embeddings using the Density-Based Spatial Clustering of Applications with Noise (DBSCAN). Then non-maximum suppression (NMS) is used to reduce redundancy. Finally, the vector representations are obtained by greedily connecting the pixels with the help of the predicted direction.
% Please see appendix for pseudocode.
% \vspace{-2mm}

% \vspace{-2mm}
\subsection{Evaluation}
\label{sec:evaluation}
% \vspace{-2mm}
In this section, we propose evaluation protocols for semantic map learning, including semantic metrics and instance metrics. % We encourage the community to do so since both information are needed in mapping problem.

% \vspace{-2mm}
\subsubsection{Semantic metrics}
\label{sec:semantic_metric}
% \vspace{-2mm}
The semantics of model predictions can be evaluated in the \textit{Eulerian} fashion and the \textit{Lagrangian} fashion. \textit{Eulerian} metrics are computed on a dense grid and measure the pixel value differences. In contrast, \textit{Lagrangian} metrics move with the shape and measure the spatial distances of shapes. 

\noindent\textbf{\textit{Eulerian} metrics.} We use intersection-over-union (IoU) as Eulerian metrics, which is given by,
\vspace{-1mm}
\begin{equation}
\label{eq:iou}
\mathrm{IoU}(\D_1, \D_2) = \frac{|\D_1\cap \D_2|}{|\D_1\cup \D_2|},
\end{equation}
% \vspace{-2mm}
where $\D_1, \D_2 \subseteq \R^{H\times W \times D}$ are dense representations of shapes (curves rasterized on a grid); $H$ and $W$ are the height and width of the grid, $D$ is number of categories; $|\cdot|$ denotes the size of the set. 

\noindent\textbf{\textit{Lagrangian} metrics.} We are interested in structured outputs, namely curves consists of connected points. To evaluate the spatial distances between the predicted curves and ground-truth curves, we use Chamfer distance (CD) of between point sets sampled on the curves:
% We evaluate the predicted curves on the HD maps in the form of point clouds. More specifically, 
% We sample points on both ground truth and predicted curves, and then use Chamfer Distance between the two point sets as the \textit{Lagrangian} metric.
\vspace{-1mm}
\begin{align}
\label{eq:chamfer_distance_directed}
&\mathrm{CD}_{\mathrm{Dir}}(\cS_1, \cS_2) = \frac{1}{\cS_1}\sum_{x\in \cS_1}\min_{y\in \cS_2} \| x-y \|_2 \\ % + \frac{1}{\cS_2}\sum_{y\in \cS_2}\min_{x\in \cS_1} \| x-y \|_2 \;\;\;
&\mathrm{CD}(\cS_1, \cS_2) = \mathrm{CD}_{\mathrm{Dir}}(\cS_1, \cS_2)+\mathrm{CD}_{\mathrm{Dir}}(\cS_2, \cS_1)
\end{align}
% \vspace{-2mm}
%and 
%\begin{equation}
%\label{eq:chamfer_distance_bidirected}
%\mathrm{CD}(\cS_1, \cS_2) = \mathrm{CD}_{\mathrm{Dir}}(\cS_1, \cS_2)+\mathrm{CD}_{\mathrm{Dir}}(\cS_2, \cS_1)
%\end{equation}
where $\mathrm{CD}_{\mathrm{dir}}$ is the directional Chamfer distance and $\mathrm{CD}$ is the bi-directional Chamfer distance; $\cS_1$ and $\cS_2$ are the two sets of points on the curves. 
% \vspace{-2mm}
\subsubsection{Instance metrics}
% \vspace{-2mm}
\label{sec:instance_metric}
We further evaluate the instance detection capability of our models. We use average precision (AP) similar to the one in object detection~\cite{lin2015microsoft}, given by
\vspace{-1mm}
\begin{equation}
    \label{eq:ap}
    \mathrm{AP} = \frac{1}{10} \sum_{r \in \{0.1, 0.2, ..., 1.0\}} \mathrm{AP}_r,
\end{equation}
where $\mathrm{AP}_r$ is the precision at recall=$r$. We collect all predictions and rank them in descending order according to the semantic confidences. Then, we classify each prediction based on the CD threshold. For example, if the CD is lower than a predefined threshold, it is considered true positive, otherwise false positive. Finally, we obtain all precision-recall pairs and compute APs accordingly.

\section{Experiments}
\label{sec:experiment}
% \vspace{-3mm}
\subsection{Implementation details}
% \vspace{-2mm}

% \textbf{Dataset.} We evaluate our approach on the NuScenes dataset \cite{caesar2020nuscenes}, which has 1000 short video sequences captured in Boston and Singapore. It includes 1,400,000 images captured with six calibrated surrounding view cameras, which leads to 360\degree coverage of the scene. Also, this dataset provides a high-quality HD Map with localization errors $\leq 10$ cm, which enables us to evaluate our online HD Map learning algorithm.

% We extract \textit{lane boundary}, \textit{lane divider}, and \textit{pedestrian crossing} as the map learning problem's prediction targets.

\noindent\textbf{Tasks \& Metrics.} We evaluate our approach on the NuScenes dataset \cite{caesar2020nuscenes}. We focus on two sub-tasks: semantic map segmentation and instance detection. Due to the limited types of map elements in the nuScenes dataset, we consider three static map elements: lane boundary, lane divider, and pedestrian crossing. 
% These three elements are the essential components in HD Maps. 
% For both tasks, we consider a bird's-eye view map. 
% The map covers [-30m, 30m] $\times$ [-15m, 15m] spatially, and the resolution is 0.15m per pixel in both directions. Therefore, the bird's-eye view grid is $200\times400$. 
% We assume the car is located at $(100, 200)$ and facing right, as shown in Figure~\ref{fig:results}. 
% To evaluate semantic segmentation, we use both IoU (Equation~\ref{eq:iou}) and CD (Equation~\ref{eq:chamfer_distance_directed}). For HD Map instance detection, We compute AP (Equation~\ref{eq:ap}) as in COCO~\cite{lin2015microsoft} at threshold CD=0.2m, 0.5m, and 1.0m.

\noindent\textbf{Architecture.} 
% Our image model consists of three parts: a perspective view image encoder, a neural view transformer, and a geometric projection module.  
For the perspective view image encoder, we adopt EfficientNet-B0~\cite{tan2020efficientnet} pre-trained on ImageNet~\cite{russakovsky2015imagenet}, as in~\cite{philion2020lift}. 
% The spatial size of the feature map is $8\times22$. 
Then, we use a multi-layer perceptron (MLP) to convert the perspective view features to bird's-eye view features in the camera coordinate system. 
% The input dimension and the output dimension of this MLP are $8\times22$ and $40\times80$, respectively. 
The MLP is shared channel-wisely and does not change the feature dimension. 
% Finally, we transform the camera coordinate features into the ego vehicle coordinate system using camera extrinsic. The features are upsampled to $200\times400$, aligned with the ground-truth map. 
% operates on each perspective-view images individually to extract the semantic information from images. The \textit{view-transformer} converts feature from image coordinate to ego coordinate and fuse the multi-view feature maps. The \textit{bev-decoder} operates on the feature maps after the multi-view fusion, aligns map elements, and outputs the semantic segmentation and instance embeddings. 
% For \textit{view-transformer}, we adopt two-layer MLP to convert the perspective features to bird's-eye view and use the provided camera extrinsic to fuse multi-camera feature maps. The input and output dim of the MLP is $H_1*W_1$ and $H_2*W_2$ where ($H_1$, $W_1$) is the shape of perspective view feature map and ($H_2$, $W_2$) is the shape of bird's-eye view feature map.
% We adopt layers from EfficientNet-B0 \cite{tan2020efficientnet} pre-trained on ImageNet \cite{russakovsky2015imagenet} for \textit{pov-encoder} and ResNet blocks for \textit{bev-encoder}, as in \cite{philion2020lift}. The \textit{bev-encoder} has two branches. The segmentation branch is trained to produce a semantic segmentation mask. In contrast, the embedding branch is optimized to output an N-dimensional embedding per pixel, such that embeddings from the same instance are close and those from different are far in the manifold, as in \cite{neven2018endtoend}.
For point clouds, we use a variant of PointPillars~\cite{lang2019pointpillars} with dynamic voxelization~\cite{Zhou2019EndtoEndMF}. We use a PointNet~\cite{Qi_2017_CVPR} with a 64-dimensional layer to aggregate points in a pillar. % After we obtain the pillar-wise features, we use convolutional three blocks with strides $[1, 2, 2]$, which gradually downsamples the input feature to 1/1, 1/2, and 1/4 of the original feature map. The three blocks have $[2, 3, 3]$ convolutional layers, with dimensions $[64, 128, 256]$. Finally, we concatenate features from these three blocks (features from the second block and the third block are upsampled to the original size). 
ResNet~\cite{resnet} with three blocks is used as the BEV decoder. 
% The final feature map is upsampled to the same size as the prediction target. 
% The three blocks have $[2, 2, 2]$ convolutional layers, with dimension $[64, 128, 256]$. 

% Specifically, we pass through the first three layers of ResNet-18, get the 3 bird's eye view features $x_1, x_2, x_3$ with different resolution, Then use two upsample layers $U_2$ and $U_4$ (where 2 and 4 is the scale factor) to get the final semantic segmentation mask $out = U_2[conv([U_4(x_3) \mathbin\Vert x_1])]$, ($\mathbin\Vert$ is concatenation) as in \cite{philion2020lift}. we use another two upsample layers $U_2'$ and $U_4'$ to get the output of instance embedding $embedding = U_2'[conv'([U_4'(x_3) \mathbin\Vert x_1])]$.

\noindent\textbf{Training details.} We use the cross-entropy loss for the semantic segmentation, and use the discriminative loss (Equation~\ref{eq:discriminative_loss}) for the instance embedding where we set $\alpha=\beta=1$, $\delta_v=0.5$, and $\delta_d=3.0$. 
%Our networks are trained for 30 epochs with batch size 4.
We use Adam \cite{kingma2017adam} for model training, witfh a learning rate of $1e-3$. %and weight decay $1e-7$. The learning rate is decayed by 0.1 for every 10 epochs. 

% \vspace{-3mm}
\subsection{Baseline methods}
\label{sec:baseline}

\begin{table*}[!tb]	
\centering
\addtolength{\tabcolsep}{0.5pt}
\caption{IoU scores (\%) and CD (m) of semantic map segmentation. CD\textsubscript{P} denotes the CD from label to prediction (equivalent to precision) while CD\textsubscript{L} denotes the CD from Prediction to Label (equivalent to recall); CD is the average of them. IoU: higher is better.  CD: lower is better. \changed{All methods use surrounding images only as input unless explicitly annotated.} *: the perspective view labels projected from 2D HD Maps. }
\label{tab:segmentation}
\footnotesize
\scalebox{0.8}{
\begin{tabular}{c|cccc|cccc|cccc|cccc}
\Xhline{2\arrayrulewidth}
\multirow{2.5}{*}{\textbf{Method}} & \multicolumn{4}{c|}{\textbf{{\scriptsize Divider}}} & \multicolumn{4}{c|}{\textbf{{\scriptsize Ped Crossing}}} & \multicolumn{4}{c|}{\textbf{{\scriptsize Boundary}}} & \multicolumn{4}{c}{\textbf{{\scriptsize All Classes}}} \\

% \cmidrule{2-5} \cmidrule{7-10} \cmidrule{12-15} \cmidrule{17-20}

& {\textbf{IoU} } & {\textbf{CD\textsubscript{P}} } & {\textbf{CD\textsubscript{L}} } & {\textbf{CD} } & {\textbf{IoU} } & {\textbf{CD\textsubscript{P}} } & {\textbf{CD\textsubscript{L}} } & {\textbf{CD} } & {\textbf{IoU} } & {\textbf{CD\textsubscript{P}} } & {\textbf{CD\textsubscript{L}} } & {\textbf{CD} } & {\textbf{IoU} } & {\textbf{CD\textsubscript{P}} } & {\textbf{CD\textsubscript{L}} } & {\textbf{CD} } \\
\hline 
IPM$^*$ & 14.4 & 1.149 & 2.232 & 2.193 & 9.5 & 1.232 & 3.432 & 2.482 & 18.4 & 1.502 & 2.569 & 1.849 & 14.1 & 1.294 & 2.744 & 2.175 \\
IPM(B) & 25.5 & 1.091 & 1.730 & 1.226 & 12.1 & 0.918 & 2.947 & 1.628 & 27.1 & 0.710 & 1.670 & 0.918 & 21.6 & 0.906 & 2.116 & 1.257 \\
IPM(CB) & 38.6 & 0.743 & 1.106 & 0.802 & 19.3 & 0.741 & 2.154 & 1.081 & 39.3 & 0.563 & 1.000 & 0.633 & 32.4 & 0.682 & 1.42 & 0.839 \\
Lift-Splat-Shoot \cite{philion2020lift} & 38.3 & 0.872 & 1.144 & 0.916 & 14.9 & 0.680 & 2.691 & 1.313 & 39.3 & 0.580 & 1.137 & 0.676 & 30.8 & 0.711 & 1.657 & 0.968 \\
VPN \cite{Pan_2020} & 36.5 & \textbf{0.534} & 1.197 & 0.919 & 15.8 & \textbf{0.491} & 2.824 & 2.245 & 35.6 & \textbf{0.283} & 1.234 & 0.848 & 29.3 & \textbf{0.436} & 1.752 & 1.337 \\
\hline \hline
HDMapNet(Surr) & 40.6 & 0.761 & 0.979 & 0.779 & 18.7 & 0.855 & 1.997 & 1.101 & 39.5 & 0.608 & 0.825 & 0.624 & 32.9 & 0.741 & 1.267 & 0.834 \\
% HDMapNet_{cam} & 0.385 & 0.720 & 1.116 & 0.793 & 0.187 & 0.697 & 2.356 & 1.128 & 0.378 & 0.659 & 0.924 & 0.700 & 0.316 & 0.692 & 1.465 & 0.874 \\
HDMapNet(LiDAR) & 26.7 & 1.134 & 1.508 & 1.219 & 17.3 & 1.038 & 2.573 & 1.524 & 44.6 & 0.501 & 0.843 & 0.561 & 29.5 & 0.891 & 1.641 & 1.101\\
HDMapNet(Fusion) & \textbf{46.1} & 0.625 & \textbf{0.893} & \textbf{0.667} & \textbf{31.4} & 0.535 & \textbf{1.715} & \textbf{0.790} & \textbf{56.0} & 0.461 & \textbf{0.443} & \textbf{0.459} & \textbf{44.5} & 0.540 & \textbf{1.017} & \textbf{0.639} \\
\Xhline{2\arrayrulewidth}
\end{tabular}
}
\normalsize
\vspace{-10pt}
\end{table*}

For all baseline methods, we use the same image encoder and decoder as HDMapNet and only change the view transformation module.

\noindent\textbf{Inverse Perspective Mapping (IPM).} The most straightforward baseline is to map segmentation predictions to the bird's-eye view via IPM~\cite{Deng_2020, 2018efficient}.
% Although IPM works well for local road layout, it has two major drawbacks: IPM assumes a flat ground plane, which is not realistic in many scenarios; 
%previews works~\cite{wang2018lanenet, Guo_2020} that focus on perspective view segmentation are not directly applicable to bird's-eye view segmentation; 

% predictions in each perspective view image are not easily ensembled to a continuous and holistic view. 

% 1. To do IPM, we need to assume the ground is a flat plane, but the ground can be very uneven in the real world, and there could be many obstructions with height on the road, like cars, pedestrians, etc. 2. Most works focus on front-view lane segmentation \cite{wang2018lanenet, Guo_2020}, which may be unsuitable for other directions.  3. If we do semantic segmentation on each image individually, identifying map instances on the HDMap is not trivial. For example, the same lane boundary may spawn in multiple cameras. Still, we don't know it from a single image before doing IPM. 
% We use FPN~\cite{lin2017feature} with 
% resnext50\_32x4d~\cite{xie2017aggregated} backbone to segment each image. Then, we project the predictions via IPM to bird's-eye view.

% We assume the ground plane is $z=0$ in the vehicle coordinate system and use intrinsic and extrinsic in nuScenes dataset to perform homographic projection.

% HD Map to image as the training labels, though the labels might be inaccurate since the localization is in 2d.

\noindent\textbf{IPM with bird's-eye view decoder (IPM(B)).} Our second baseline is an extension of IPM. Rather than making predictions in perspective view, we perform semantic segmentation directly in bird-eye view. % To consider the surrounding cameras' information as a whole, we can firstly warp perspective-view images to birds' eye view and then perform semantic segmentation directly from the topdown view. 
% We use the same decoder network as in HDMapNet.

\noindent\textbf{IPM with perspective view feature encoder and bird's-eye view decoder (IPM(CB)).} The next extension is to perform feature learning in the perspective view while making predictions in the bird's-eye view.
% We use the same image encoder and BEV decoder as in HDMapNet.
% A pipeline more similar to us is to firstly extract features from perspective-view, fusion them by a homography, and then decode the feature map by a bird's eye view decoder. We use the same encoder and decoder as in HDMapNet.

\noindent\textbf{Lift-Splat-Shoot.} Lift-Splat-Shoot \cite{philion2020lift} estimates a distribution over depth in the perspective view images. Then, it converts 2D images into 3D point clouds with features and projects them into the ego vehicle frame.
% Finally, the point clouds are transformed to a bird's-eye view feature map with PointPillars~\cite{lang2019pointpillars}. %then follow the pointpillars  \cite{lang2019pointpillars}architecture to convert the point cloud into a top-down feature map. 
% We use the code provided by \cite{philion2020lift} for experiments.

\noindent\textbf{View Parsing Network (VPN).} VPN~\cite{Pan_2020} proposes a simple view transformation module: a view relation module to model the relations between any two pixels and a view fusion module to fuses the features of pixels.  % fuse them by a \textit{View Fusion Module (VFM)}. 
% , Two-layer MLPs with the fixed resolution are used for \textit{VRM} and Avg-pooling layer for \textit{VFM}.
% We use the code released by \cite{Pan_2020} for experiments.

% \begin{table*}[tbp]
% \centering
% \addtolength{\tabcolsep}{3.5pt}
% \caption{The results of instance map detection}
% \label{tab:detection}
% \footnotesize
% \begin{tabular}{l|lccccccc}
% \Xhline{2\arrayrulewidth}
% \textbf{Method} & \textbf{Class} & \textbf{mAP@0.2m} & \textbf{mAP@0.5m} & \textbf{mAP@1.0m} & \textbf{mAP}  \\ 
% \hline 
% \multirow{3}{*}{IPM + bev}
% & Divider & \\
% & Ped Crossing & \\
% & Boundary & \\
% \hline \hline
% \multirow{3}{*}{cam + IPM + bev}
% & Divider & 0.102 & 0.250 & 0.368 \\
% & Ped Crossing & 0.020 & 0.078 & 0.122 \\
% & Boundary & 0.101 & 0.279 & 0.455 \\
% \hline \hline
% \multirow{3}{*}{Lift-Splat-Shoot}
% & Divider & 0.091 & 0.238 & 0.359 \\
% & Ped Crossing & 0.009 & 0.054 & 0.089 \\
% & Boundary & 0.085 & 0.229 & 0.412 \\
% \hline \hline
% \multirow{3}{*}{VPN}
% & Divider & \textbf{0.137} & 0.307 & \textbf{0.406} \\
% & Ped Crossing & 0.019 & 0.074 & 0.121 \\
% & Boundary & \textbf{0.137} & 0.339 & 0.501 \\
% \hline \hline
% \multirow{3}{*}{HDMapNet (camera)}
% & Divider & 0.112 & 0.269 & 0.386 \\
% & Ped Crossing & 0.021 & 0.079 & 0.133 \\
% & Boundary & 0.115 & 0.324 & 0.498 \\
% % \multirow{3}{*}{VPN + IPM}
% % & Divider & 0.108 & 0.268 & 0.383\\
% % & Ped Crossing & \textbf{0.033} & \textbf{0.088} & \textbf{0.135}\\
% % & Boundary & 0.130 & \textbf{0.349} & \textbf{0.518} \\
% \Xhline{2\arrayrulewidth}
% \end{tabular}
% \normalsize
% \vspace{-5pt}
% \end{table*}

\begin{table*}[!tb]	
\centering
% \addtolength{\tabcolsep}{0.5pt}
\caption{Instance detection results. \{0.2, 0.5, 1.\} are the predefined thresholds of Chamfer distance (\eg a prediction is considered a true positive if the Chamfer distance between label and prediction is lower than that threshold). The mAP is the average of three APs. AP \& mAP: higher is better. \changed{All methods only use surrounding images as input unless explicitly annotated.}}
\label{tab:detection}
\footnotesize
\scalebox{0.8}{
\begin{tabular}{c|cccc|cccc|cccc|cccc}
\Xhline{2\arrayrulewidth}
\multirow{2.5}{*}{\textbf{Method}} & \multicolumn{4}{c|}{\textbf{{\scriptsize Divider}}} & \multicolumn{4}{c|}{\textbf{{\scriptsize Ped Crossing}}} & \multicolumn{4}{c|}{\textbf{{\scriptsize Boundary}}} & \multicolumn{4}{c}{\textbf{{\scriptsize All Classes}}} \\

% \cmidrule{2-5} \cmidrule{7-10} \cmidrule{12-15} \cmidrule{17-20}

& {\textbf{AP@.2}} & {\textbf{AP@.5}} & {\textbf{AP@1.}} & {\textbf{mAP}} & {\textbf{AP@.2}} & {\textbf{AP@.5}} & {\textbf{AP@1.}} & {\textbf{mAP}} & {\textbf{AP@.2}} & {\textbf{AP@.5}} & {\textbf{AP@1.}} & {\textbf{mAP}} & {\textbf{AP@.2}} & {\textbf{AP@.5}} & {\textbf{AP@1.}} & {\textbf{mAP}}  \\
\hline 
IPM(B) & 2.6 & 9.8 & 19.6 & 10.7 & 1.6 & 4.8 & 7.8 & 4.7 & 2.2 & 9.2 & 23.7 & 11.7 & 2.1 & 7.9 & 17.0 & 9.0 \\
IPM(CB) & 10.2 & 25.0 & 36.8 & 24.0 & 2.0 & 7.8 & 12.2 & 7.3 & 10.1 & 27.9 & 45.5 & 27.8 & 7.4 & 20.2 & 31.5 & 19.7 \\
Lift-Splat-Shoot \cite{philion2020lift} & 9.1 & 23.8 & 35.9 & 22.9 & 0.9 & 5.4 & 8.9 & 5.1 & 8.5 & 22.9 & 41.2 & 24.2 & 6.2 & 17.4 & 28.7 & 17.4\\
VPN \cite{Pan_2020} & 8.8 & 22.7 & 34.9 & 22.1 & 1.2 & 5.3 & 9.0 & 5.2 & 9.2 & 24.1 & 42.7 & 25.3 & 6.4 & 17.4 & 28.9 & 17.5  \\
\hline \hline
HDMapNet(Surr) & 13.7 & 30.7 & 40.6 & 28.3 & 1.9 & 7.4 & 12.1 & 7.1 & 13.7 & 33.9 & 50.1 & 32.6 & 9.8 & 24.0 & 34.3 & 22.7 \\
HDMapNet(LiDAR) & 1.0 & 5.7 & 15.1 & 7.3 & 1.9 & 5.8 & 9.0 & 5.6 & 5.8 & 20.2 & 39.6 & 21.9 & 2.9 & 10.6 & 21.2 & 11.6 \\
HDMapNet(Fusion) & \textbf{15.0} & \textbf{32.6} & \textbf{46.0} & \textbf{31.2} & \textbf{6.4} & \textbf{13.6} & \textbf{17.8} & \textbf{12.6} & \textbf{26.7} & \textbf{51.5} & \textbf{65.6} & \textbf{47.9} & \textbf{16.0} & \textbf{32.6} & \textbf{43.1} & \textbf{30.6} \\
\Xhline{2\arrayrulewidth}
\end{tabular}
}
\normalsize
\vspace{-10pt}
\end{table*}

\begin{figure*}[!t]
\centering
\includegraphics[width=0.23\textwidth]{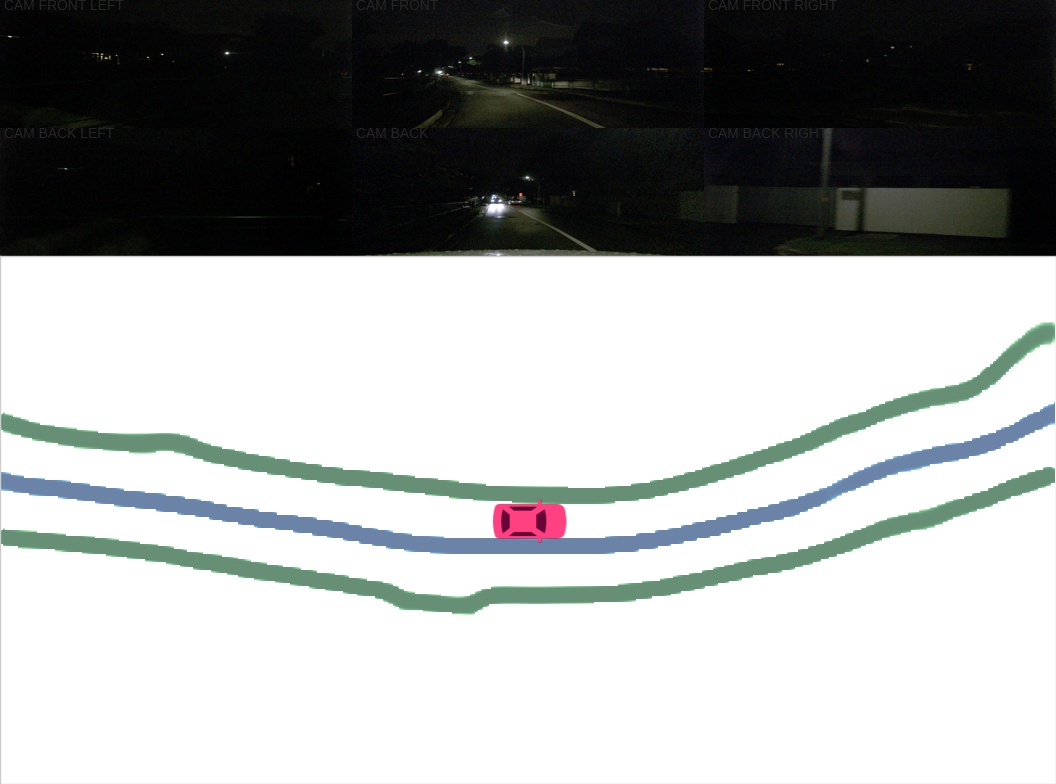}
\includegraphics[width=0.23\textwidth]{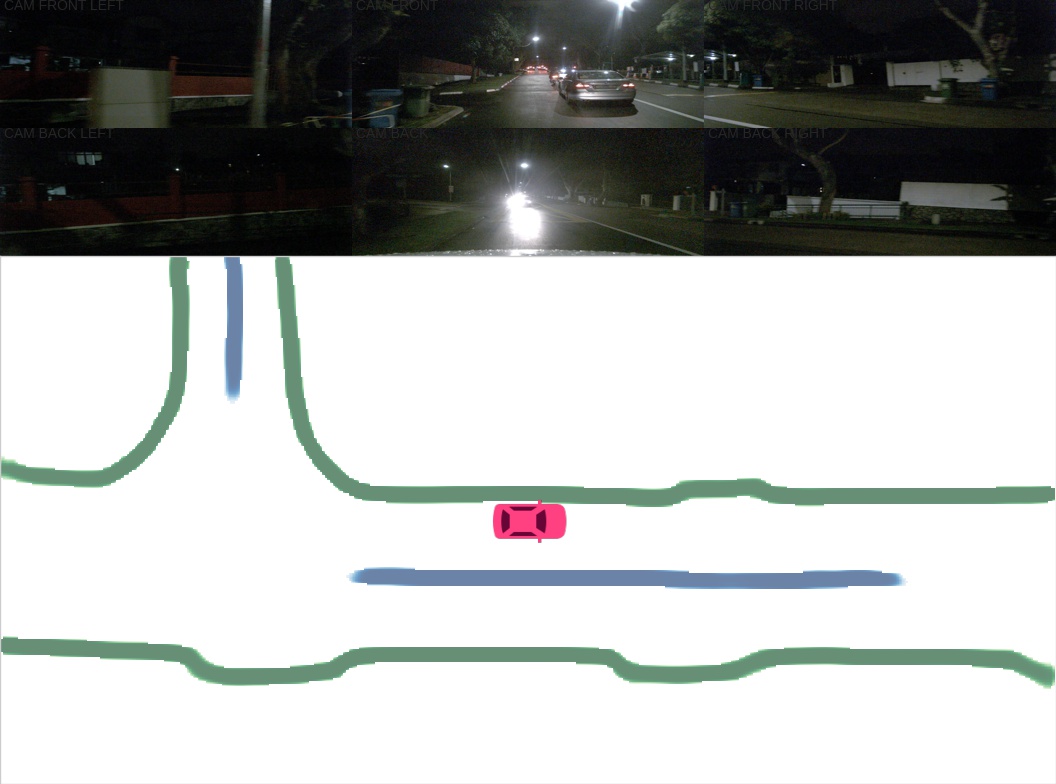}
\includegraphics[width=0.23\textwidth]{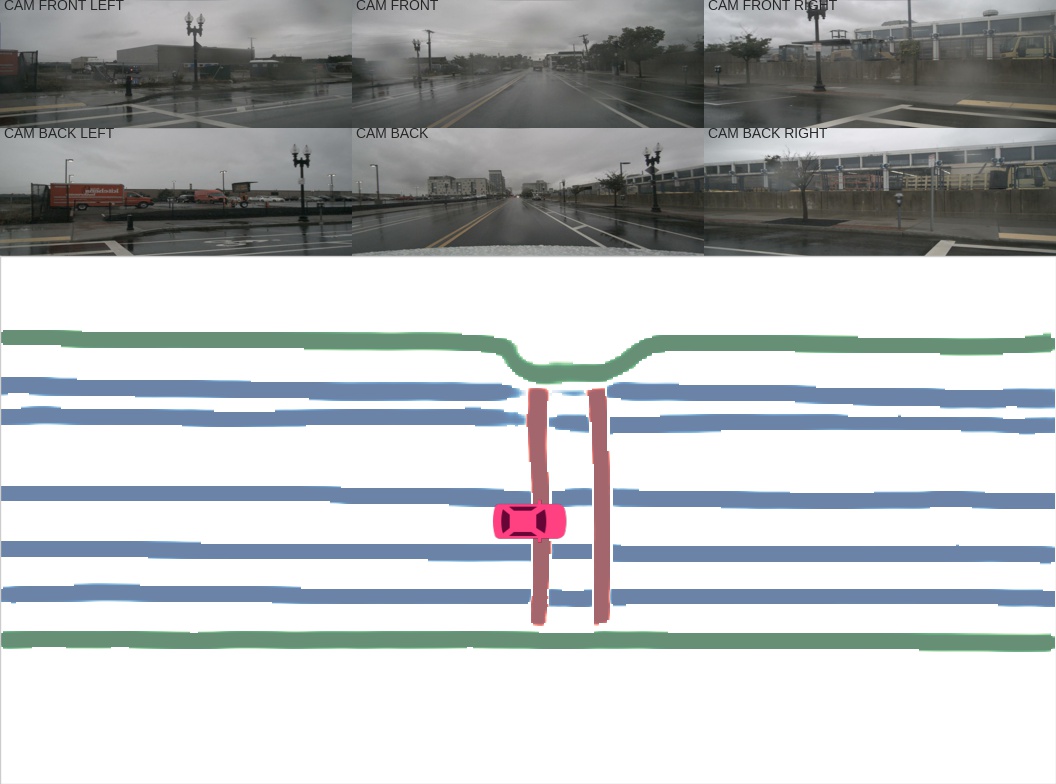}
\includegraphics[width=0.23\textwidth]{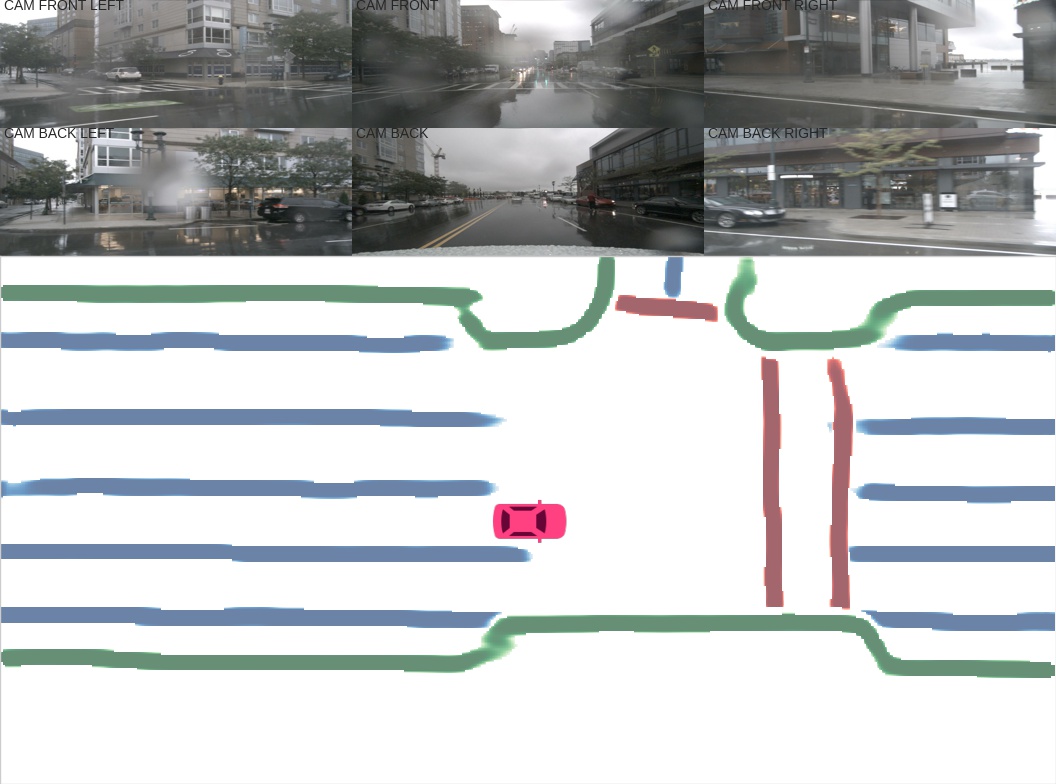}
\caption{{Qualitative results under bad weather conditions. The left two images show predictions at night. The right twos show predictions at rainy days.} \label{fig:weather}}
\vspace{-2em}

\end{figure*}

% \vspace{-4mm}
\subsection{Results}
% \vspace{-2mm}

%\textbf{Semantic map segmentation.}
We compare our HDMapNet against baselines  in Section~\ref{sec:baseline}. Table~\ref{tab:segmentation} shows the comparisons. First, Our HDMapNet(Surr), which is the surrounding camera-only method, outperforms all baselines. This suggests that our novel learning-based view transformation is indeed effective, without making impractical assumptions about a complex ground plane (IPM) or estimating the depth (Lift-Splat-Shoot). Second, our HDMapNet(LiDAR) is better than HDMapNet(Surr) in boundary but worse in divider and pedestrian crossing. This indicates different categories are not equally recognizable in one modality. Third, our fusion model with both camera images and LiDAR point clouds achieves the best performance. It improves over baselines and our camera-only method by 50\% relatively. 
% we can see our HDMapNet(Fusion) outperforms all other methods by a large margin. Even for the camera-only methods, our HDMapNet(Surr) is also competitive. This may indicate the learning-based view transformation is a simple yet strong baseline, without the need to make impractical assumptions or build a complex plane model, like in IPM and Lift-Splat-Shoot. Note that though both HDMapNet and VPN use fully-connected layers to do view transformation, our method has a clear improvement. We suggest this improvement comes from the higher resolution given by the variable output size of the transformation layer and the better fusion strategy by using geometric projection.

Another interesting phenomenon is that various models behave differently in terms of the CD. For example, VPN has the lowest CD\textsubscript{P} in all categories, while it underperforms its counterparts on CD\textsubscript{L} and has the worst overall CD. Instead, our HDMapNet(Surr) balances both CD\textsubscript{P} and CD\textsubscript{L}, achieving the best CD among all camera-only-based methods. This finding indicates that CD is complementary to IoU, which shows the precision and recall aspects of models. This helps us understand the behaviors of different models from another perspective.

% \begin{figure}[t!]
%     \centering
%     \begin{subfigure}[t]{.32\textwidth}
%         \centering
%         \includegraphics[width=0.9\linewidth]{section/figs/1_frame.png}
%         % \vspace{-0.5em}
%         \caption*{1 frame}
%     \end{subfigure}
%     \begin{subfigure}[t]{.32\textwidth}
%         \centering
%         \includegraphics[width=0.9\linewidth]{section/figs/8_frame.png}
%         % \vspace{-0.5em}
%         \caption*{8 frames}
%     \end{subfigure}
%     \begin{subfigure}[t]{.32\textwidth}
%         \centering
%         \includegraphics[width=0.9\linewidth]{section/figs/20_frame.png}
%         % \vspace{-0.5em}
%         \caption*{20 frames}
%     \end{subfigure}
%     \caption{}
%     % \vspace{-6mm}
%     \label{fig:results}
% \end{figure}

\begin{figure}[!t]
\centering
\includegraphics[width=0.5\textwidth]{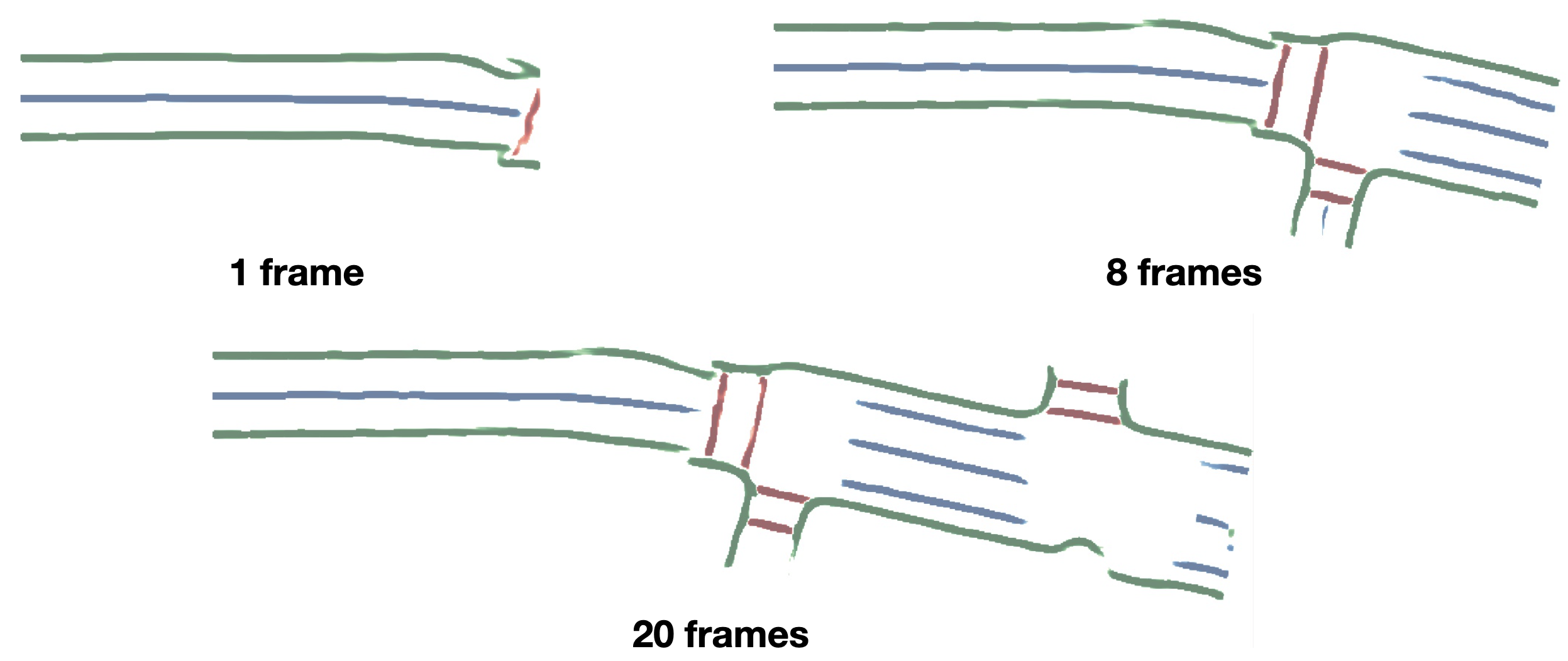}
\caption{Long-term temporal accumulation by fusing occupancy probabilities over multiple frames. \label{fig:temporal}}
\vspace{-2em}

\end{figure}

\noindent\textbf{Instance map detection.}
In Figure~\ref{fig:overview} (Instance detection branch), we show the visualization of embeddings using principal component analysis (PCA). Different lanes are assigned different colors even when they are close to each other or have intersections. This confirms our model learns instance-level information and can predict instance labels accurately. 
In Figure~\ref{fig:overview} (Direction classification branch), we show the direction mask predicted by our direction branch. The direction is consistent and smooth.
% It can be seen that different lanes are assigned a different color, even when two lanes are closed to each other or have intersections, which shows the instance semantic learned by embedding layers. 
We show the vectorized curve produced after post processing in Figure~\ref{fig:results}.
% Finally, we cluster pixels of the same instance ID, perform NMS, and connect pixels by direction to get a vectorized curve representation as shown in Figure~\ref{fig:results}.
% After clustering and post-process, we can get a vectorized local HD Map. See Figure~\ref{fig:results} for visualization. 
In Table~\ref{tab:detection}, we present the quantitative results of instance map detection. HDMapNet(Surr) already outperforms baselines while HDMapNet(Fusion) is significantly better than all counterparts, e.g., it improves over IPM by 55.4\%.

\noindent\textbf{Sensor fusion.}
In this section, we further analyze the effect of sensor fusion for constructing HD semantic maps. As shown in Table~\ref{tab:segmentation}, for divider and pedestrian crossing, HDMapNet(Surr) outperforms HDMapNet(LiDAR), while for lane boundary, HDMapNet(LiDAR) works better. We hypothesize this is because there are elevation changes near the lane boundary, making it easy to detect in LiDAR point clouds. On the other hand, the color contrast of road divider and pedestrian crossing is helpful information, making two categories more recognizable in images; visualizations also confirm this in Figure~\ref{fig:results}. The strongest performance is achieved when combining LiDAR and cameras; the combined model outperforms both models with a single sensor by a large margin. This suggests these two sensors include complementary information for each other.% and leave a better fusion strategy for future work.,

{
\noindent\textbf{Bad weather conditions.}
Here we assess the robustness of our model under extreme weather conditions. As shown in Figure~\ref{fig:weather}, our model can generate complete lanes even when lighting condition is bad, or when the rain obscures sight. We speculate that the model can predict the shape of the lane based on partial observations when the roads are not completely visible. Although there are performance drop in extreme weather condition, the overall performance is still reasonable. (Table~\ref{tab:weather})

\noindent\textbf{Temporal Fusion.}
Here we experiment on temporal fusion strategies. We first conduct short-term temporal fusion by pasting feature maps of previous frames into current's according to ego poses. The feature maps are fused by max pooling and then fed into decoder. As shown in Table~\ref{tab:temporal_fusion}, fusing multiple frames can improve the IoU of the semantics.
We further experiment on long-term temporal accumulation by fusing segmentation probabilities. As shown in Figure~\ref{fig:temporal}, our method produces consistent semantic maps with larger field of view while fusing multiple frames.
}

\begin{table}[t]	
\centering
\addtolength{\tabcolsep}{2.5pt}
\begin{tabular}{c|ccc}
\Xhline{2\arrayrulewidth}
\textbf{Weather} & Night & Rainy & Normal \\ \hline
\textbf{IoU}  & 39.3 & 38.7 & 44.9 \\
\Xhline{2\arrayrulewidth}
\end{tabular}
\caption{The semantic segmentation performance of HDMapNet(Fusion) under different weather.}
\label{tab:weather}
\end{table}

\begin{table}[t]	
\centering
\addtolength{\tabcolsep}{2.5pt}
\begin{tabular}{c|ccc}
\Xhline{2\arrayrulewidth}
\textbf{\# of frames} & 1 & 2 & 4 \\ \hline
\textbf{IoU}  & 32.9 & 35.8 & 36.4 \\
\Xhline{2\arrayrulewidth}
\end{tabular}
\caption{Temporal fusion on HDMapNet(Surr). 1 means temporal fusion is not used.}
\label{tab:temporal_fusion}
\vspace{-2em}
\end{table}

% \begin{minipage}{\textwidth}
%  \begin{minipage}[t]{0.48\textwidth}
%   \centering
%         \makeatletter\def\@captype{table}\makeatother\caption{{The semantic segmentation performance of HDMapNet(Fusion) under different weather.}}
%         \begin{tabular}{c|ccc}
%         \Xhline{2\arrayrulewidth}
%         \textbf{Weather} & Night & Rainy & Normal \\ \hline
%         \textbf{IoU}  & 39.3 & 38.7 & 44.9 \\
%         \Xhline{2\arrayrulewidth}
%         \end{tabular}
%         \label{tab:weather}
%   \end{minipage}
%   \begin{minipage}[t]{0.48\textwidth}
%   \centering
  
%         \makeatletter\def\@captype{table}\makeatother\caption{{Temporal fusion on HDMapNet(Surr). 1 means temporal fusion is not used.}}
%         \begin{tabular}{c|ccc}
%         \Xhline{2\arrayrulewidth}
%         \textbf{\# of frames} & 1 & 2 & 4 \\ \hline
%         \textbf{IoU}  & 32.9 & 35.8 & 36.4 \\
%         \Xhline{2\arrayrulewidth}
%         \end{tabular}
%         \label{tab:temporal_fusion}
%   \end{minipage}
% \end{minipage}
\section{Conclusion}
HDMapNet predicts HD semantic maps directly from camera images and/or LiDAR point clouds.
The local semantic map learning framework could be a more scalable approach than the {global map construction} and annotation pipeline that requires a significant amount of human efforts. 
{Even though our baseline method of semantic map learning does not produce map elements as accurate, it gives system developers another possible choice of the trade-off between scalability and accuracy. }

%We hope our method provides insights into design choices of the perception and the mapping in autonomous driving and robotics broadly. 

% Beyond its direct usage, our experiments suggest several avenues for future inquiry. First, camera images and LiDAR point clouds complement each other; adding more modalities and searching for better fusion strategies can potentially improve the generalizability and stability of the self-driving system. 
% Additionally, using the online maps for downstream tasks such as behavior prediction, motion planning remains an interesting and open problem. 

\newpage

\bibliographystyle{IEEEtran}
\bibliography{IEEEabrv, IEEEexample}

%%%%%%%%%%%%%%%%%%%%%%%%%%%%%%%%%%%%%%%%%%%%%%%%%%%%%%%%%%%%%%%%%%%%%%%%%%%%%%%%

\addtolength{\textheight}{-12cm}   % This command serves to balance the column lengths

%%%%%%%%%%%%%%%%%%%%%%%%%%%%%%%%%%%%%%%%%%%%%%%%%%%%%%%%%%%%%%%%%%%%%%%%%%%%%%%%

%%%%%%%%%%%%%%%%%%%%%%%%%%%%%%%%%%%%%%%%%%%%%%%%%%%%%%%%%%%%%%%%%%%%%%%%%%%%%%%%

%%%%%%%%%%%%%%%%%%%%%%%%%%%%%%%%%%%%%%%%%%%%%%%%%%%%%%%%%%%%%%%%%%%%%%%%%%%%%%%%

%%%%%%%%%%%%%%%%%%%%%%%%%%%%%%%%%%%%%%%%%%%%%%%%%%%%%%%%%%%%%%%%%%%%%%%%%%%%%%%%

\end{document}